\documentclass[10pt,twocolumn,letterpaper]{article}

\usepackage[pagenumbers]{cvpr} 

\definecolor{cvprblue}{rgb}{0.21,0.49,0.74}
\usepackage[pagebackref,breaklinks,colorlinks,allcolors=cvprblue]{hyperref}

\usepackage{multirow}
\usepackage{algorithm}
\usepackage{algpseudocode}
\usepackage{graphicx}
\usepackage{mathtools}
\usepackage{hyperref}

\newcommand*{\affaddr}[1]{#1} 
\newcommand*{\affmark}[1][*]{\textsuperscript{#1}}
\def\paperID{} 
\def\confName{CVPR}
\def\confYear{2026}

\title{PropFly: Learning to Propagate via On-the-Fly Supervision \\from Pre-trained Video Diffusion Models}

\author{
Wonyong Seo\affmark[1]\footnotemark[1] \quad 
Jaeho Moon\affmark[1]\footnotemark[1] \quad 
Jaehyup Lee\affmark[2]\footnotemark[2] \quad 
Soo Ye Kim\affmark[3]\footnotemark[2] \quad 
Munchurl Kim\affmark[1]\footnotemark[2]\\
\affaddr{\affmark[1]KAIST} \quad
\affaddr{\affmark[2]Kyungpook National University} \quad
\affaddr{\affmark[3]Adobe Research}\\
\affaddr{\small{\url{https://kaist-viclab.github.io/PropFly_site/}}}
}

\begin{document}

\twocolumn[{%
\renewcommand\twocolumn[1][]{#1}%
\maketitle
\vspace{-8mm}
\begin{center}\centering
    \includegraphics[width=0.95\textwidth]{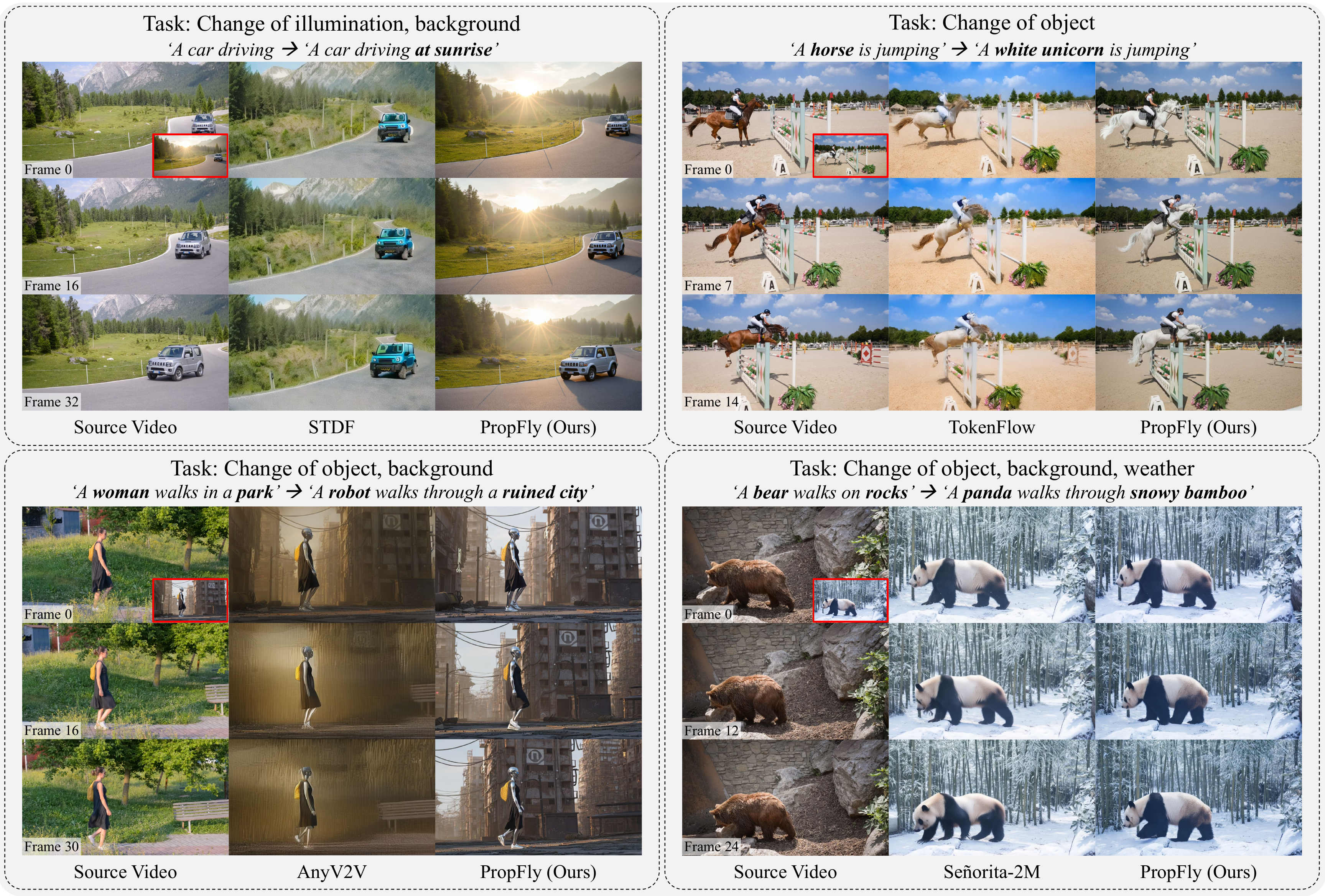}
    \captionof{figure}{Qualitative comparison of our PropFly against \textbf{text-guided} (STDF \cite{yatim2024space_stdf}, TokenFlow \cite{tokenflow2023}) and \textbf{propagation-based} (AnyV2V \cite{ku2024anyv2v}, Señorita-2M~\cite{zi2025senorita}) video editing methods. Our PropFly demonstrates robust performance across a wide range of edits, from local editing to complex transformations. Note that all propagation-based methods were conditioned on the \textbf{same edited frames (in red boxes).}}
   \label{fig:figure1}
\end{center}
}]

\begin{abstract}
Propagation-based video editing enables precise user control by propagating a single edited frame into following frames while maintaining the original context such as motion and structures.
However, training such models requires large-scale, paired (source and edited) video datasets, which are costly and complex to acquire.
Hence, we propose the \textbf{PropFly}, a training pipeline for \textbf{Prop}agation-based video editing, relying on on-the-\textbf{Fly} supervision from pre-trained video diffusion models (VDMs) instead of requiring off-the-shelf or precomputed paired video editing datasets.
Specifically, our PropFly leverages one-step clean latent estimations from intermediate noised latents with varying Classifier-Free Guidance (CFG) scales to synthesize diverse pairs of `source' (low-CFG) and `edited' (high-CFG) latents on-the-fly. 
The source latent serves as structural information of the video, while the edited latent provides the target transformation for learning propagation.
Our pipeline enables an additional adapter attached to the pre-trained VDM to learn to propagate edits via Guidance-Modulated Flow Matching (GMFM) loss, which guides the model to replicate the target transformation.
Our on-the-fly supervision ensures the model to learn temporally consistent and dynamic transformations.
Extensive experiments demonstrate that our PropFly significantly outperforms the state-of-the-art methods on various video editing tasks, producing high-quality editing results.

\end{abstract}    

{
  \renewcommand{\thefootnote}%
    {\fnsymbol{footnote}}
  \footnotetext[1]{Co-first authors (equal contribution).}
  \footnotetext[2]{Co-corresponding authors.}
}

\section{Introduction}
\label{sec:introduction}


The advent of powerful generative models, such as diffusion-based models~\cite{ho2020denoising, rombach2022high, peebles2023scalable_dit,liu2022flow, lipman2022flow}, has enabled unprecedented realism in visual synthesis. 
This success is now extending to the video domain~\cite{ho2022video, singer2022make, ho2022imagenvideo, blattmann2023stable, esser2023structure, yang2024cogvideox, hacohen2024ltx, wan2025wan_lucy, kong2412hunyuanvideo, polyak2024movie}, offering powerful tools to automate and simplify complex video editing tasks.
The dominant paradigm for such video editing methods is a text-conditional approach~\cite{wu2023tune, tokenflow2023, yatim2024space_stdf, cheng2023consistent_insv2v, singer2024video_eve}, which offers an intuitive user experience. 
These models possess remarkable generative capabilities, enabling them to synthesize changes (e.g., style transfer or local object manipulation) guided by text. 
However, in practice, it is challenging to describe the exact, fine-grained visual attributes of desired edits, often leading to results that do not perfectly reflect the user's creative intent.


The inherent limitations of text-based control have motivated propagation-based video editing \cite{ku2024anyv2v, feng2024ccedit, liu2025generative}, which offers more controllability by propagating a precisely edited single frame to the entire video. 
However, training such models can be challenging due to the scarcity of large-scale diverse paired video datasets (i.e., source and edited videos). 
To circumvent this, GenProp~\cite{liu2025generative} synthesizes training pairs based on object segmentation masks, but this approach is only tailored for local changes such as object addition and removal, and cannot generate data pairs for global transformations like artistic stylization.
Other approaches~\cite{feng2024ccedit, burgert2025gowiththeflow} rely on auxiliary guidance signals such as pre-computed depth maps or optical flows to avoid using paired data. 
This dependency, however, makes them highly susceptible to artifacts stemming from inaccuracies in the guidance signals.
While Señorita-2M~\cite{zi2025senorita} employs recent diffusion-based models to synthesize paired training datasets, this approach can be computationally expensive, especially for videos, due to the iterative diffusion inference process.
Moreover, their data pipeline only supports a limited range of video editing tasks.

To address such limitations, we propose PropFly, \textbf{Prop}agation-based video editing training pipeline via on-the-\textbf{Fly} supervision from pre-trained video diffusion models (VDMs), without requiring any off-the-shelf or precomputed paired video datasets.
Our key insight is to use the pre-trained VDM's generative capability as the source of supervision by exploiting varying Classifier-Free Guidance (CFG)~\cite{ho2022classifier_cfg} scales to generate video latent pairs for training.
Such pairs are structurally aligned but semantically distinct and thus can serve as the source of supervision by learning the transformation between them.
This data generation process is made computationally efficient by leveraging a \textit{one-step} clean latent estimation from intermediate-noised video latents instead of running the full iterative diffusion sampling process.
A trainable adapter, attached to the pre-trained VDM for propagating the edits, is then conditioned on the entire \textit{source} latent frames and the first frame of \textit{edited} latents from the data pairs generated on the fly.
The adapter is trained via a Guidance-Modulated Flow Matching (GMFM) loss to apply the target transformation to the source based on the edited latent.
To further enrich this supervision signal, we apply Random Style Prompt Fusion (RSPF) to generate diverse training examples.

PropFly shows improved propagation-based video editing quality across a wide range of video editing tasks, from local edits to global transformations.
As shown in Fig.~\ref{fig:figure1}, our PropFly overcomes the limitations of text-guided methods like STDF~\cite{yatim2024space_stdf} and TokenFlow~\cite{tokenflow2023}, which often struggle to apply edits precisely while preserving the video's original content.
Also, PropFly shows superior fidelity in complex transformations compared to other propagation-based methods, including AnyV2V~\cite{ku2024anyv2v} and Señorita-2M~\cite{zi2025senorita}.
Our Propfly achieves significantly improved state-of-the-art (SOTA) performance on recent video editing benchmarks~\cite{ju2025editverse, wu2023tgvecvpr} in terms of video quality, text alignment, and temporal consistency. 
Our key contributions are summarized as follows:

\begin{itemize}
    \item We propose PropFly, a novel training pipeline for propagation-based video editing using the video generation capability of pre-trained VDMs, without requiring paired video datasets or auxiliary guidance signals.

    \item Based on our CFG modulation with one-step clean latent estimation, pre-trained VDMs can generate on-the-fly supervision signals in a computationally efficient manner, enabling our adapter to propagate various video edits.

    \item Our novel GMFM loss can effectively guide the model to learn the transformation between the on-the-fly data pairs.

    \item Our PropFly shows remarkable global video edit propagation quality and achieves significantly improved performance on recent video editing benchmarks.
\end{itemize}

\begin{figure*}[t!]
    \centering
    \includegraphics[width=\linewidth]{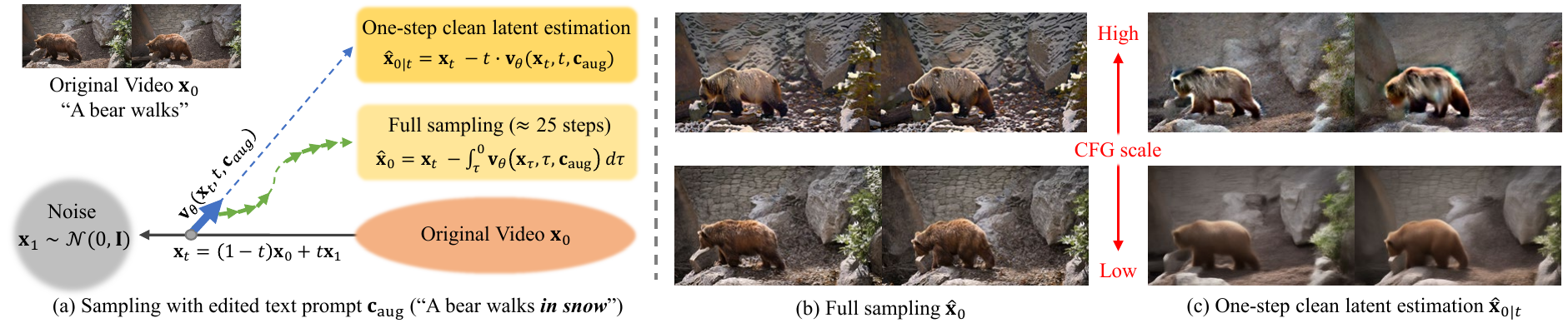}
    \caption{
    An illustration of our on-the-fly data pair generation process based on one-step clean latent estimation.
    (a) Pre-trained VDM sampling process from intermediate noised latents \(\mathbf{x}_t\) with an edited text prompt \(\mathbf{c}_\text{aug}\), showing clean latent estimation after one-step sampling (Eq.~\ref{eq:onestep}) and full sampling (an iterative ODE solve from $t$ to $0$).
    (b) Increasing the CFG scale ($\omega$) progressively strengthens the semantic edit (i.e., altering style, texture, and color).
    (c) Our method leverages this phenomenon efficiently: instead of performing computationally expensive full sampling, we utilize one-step clean latent predictions generated at a low CFG scale ($\omega_L$) and a high CFG scale ($\omega_H$). 
    These on-the-fly predictions serve as the aligned source ($\hat{\mathbf{x}}_{0|t}^{\text{low}}$) and target ($\hat{\mathbf{x}}_{0|t}^{\text{high}}$) pair for training our PropFly.}
    \label{fig:onestep}
    \vspace{-2mm}
\end{figure*}

\section{Related Work}
\label{sec:related_work}



\subsection{Text-guided Video Editing}

Text instruction-based video editing models aim to modify a source video following a user-provided text prompt.
Many methods adapt foundational image editing concepts to the video domain such as cross-attention manipulation from Prompt-to-Prompt~\cite{hertz2022prompt}. 
These approaches generally fall into two categories.
The first category includes training-free methods that typically propagate diffusion features~\cite{tokenflow2023, qi2023fatezero, Ceylan_2023_ICCV, yatim2024space_stdf} or manipulate attention maps~\cite{Kara_2024_CVPR, shin2024edit, wang2023zero, wu2024fairy} to maintain inter-frame correspondence during synthesis.
However, training-free methods often rely on per-video optimization or DDIM inversion~\cite{mokady2023null} processes that incur extensive inference time and suffer from inconsistent performance depending on the input videos.
The second category involves training or fine-tuning. 
Some methods require per-video fine-tuning on the input video to adapt to a new subject or style~\cite{wu2023tune, bar2022text2live, Liu_2024_CVPR, zhao2025controlvideo}. 
Others~\cite{singer2024video_eve, Jeong_2024_CVPR_vmc, zhao2024motiondirector} train dedicated adapters on video datasets, which can then be applied to new videos. 
InsV2V~\cite{cheng2023consistent_insv2v} involves training a general-purpose video-to-video translation model on a large-scale synthetic dataset.
However, text-based video editing methods often struggle to reflect user intent, especially when making fine-grained edits or applying a global artistic style. 

\subsection{Propagation-based Video Editing}

To overcome the limitations of text-based control, another line of research focuses on propagating edits from a single frame throughout the video, from local object manipulation~\cite{liu2025generative, mou2024revideo, gu2024videoswap} to the challenging tasks of global video editing~\cite{ku2024anyv2v,feng2024ccedit}, such as artistic stylization, and weather or lighting changes. 
AnyV2V~\cite{ku2024anyv2v} leverages an inversion-based approach for propagation-based video editing, while I2VEdit~\cite{ouyang2024i2vedit} introduces per-video test-time optimization using an I2V model.
However, these approaches introduce significant computational overhead at inference time, limiting their practicality.
Other methods, such as CCEdit~\cite{feng2024ccedit} and Go-with-the-Flow \cite{burgert2025gowiththeflow}, rely on auxiliary information (e.g., optical flows or depth maps) to preserve the source video's motion and structure during editing instead of directly conditioning on the source RGB frames. 
As a result, their performance becomes highly sensitive to the errors in these guidance signals, making them susceptible to artifacts.
Recently, Genprop~\cite{liu2025generative} proposed data augmentation techniques to synthesize training data using object masks, which is effective for local edits such as object addition or removal but cannot handle global transformations like holistic appearance changes. 
In contrast, our PropFly trains propagation-based video editing models using \textit{on-the-fly} supervision from frozen pre-trained VDMs.
By generating structurally aligned yet semantically diverse latent pairs during training, PropFly provides rich and flexible supervision and further supports global edits without relying on explicit paired datasets or auxiliary guidance signals.

\begin{figure*}[t!]
    \centering
    \includegraphics[width=\linewidth] {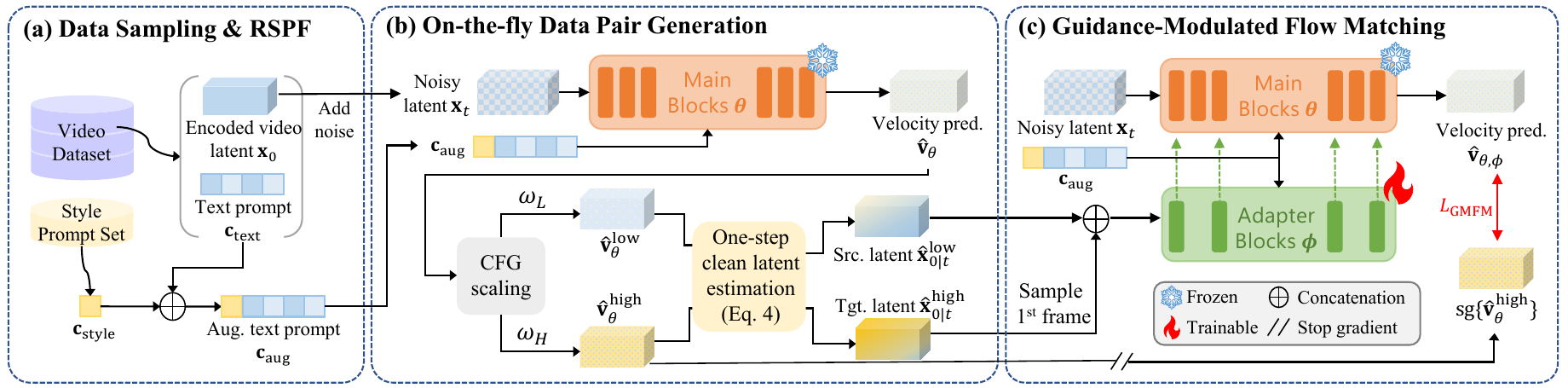}
    \caption{Overview of our PropFly training pipeline. 
    (a) A pair of video \(\mathbf{x}_0\) and text prompt \(\mathbf{c}_\text{text}\) is sampled from the video dataset and an augmented text \(\mathbf{c}_\text{aug}\) is synthesized, by appending random style prompt \(\mathbf{c}_\text{style}\) to \(\mathbf{c}_\text{text}\). 
    (b) A frozen, pre-trained VDM \(\theta\) synthesizes a data pair (\(\hat{\mathbf{x}}_{0|t}^{\text{low}}, \hat{\mathbf{x}}_{0|t}^{\text{high}}\)) on the fly from a single noised latent \(\mathbf{x}_t\) using low and high CFG scales (guided by \(\mathbf{c}_\text{aug}\)). 
    (c) A trainable adapter \(\phi\) with the frozen VDM \(\theta\) is then conditioned on the source video latent \(\hat{\mathbf{x}}_{0|t}^{\text{low}}\) (for structure) and the edited first frame latent of \(\hat{\mathbf{x}}_{0|t}^{\text{high}}\). 
    The adapter is trained via GMFM loss to predict the VDM's text-guided, high-CFG velocity, effectively learning to edit the remaining video frames.}
    \label{fig:main_overview}
    \vspace{-2mm}
\end{figure*}

\subsection{Training data for Video Editing}
The quality of generative video editing models is heavily dependent on large-scale, high-quality training datasets.
To address data scarcity, several approaches bring state-of-the-art editing diffusion models to their training data pipeline as a form of data augmentation~\cite{alimisis2025advances, kawar2023imagic, kim2022diffusionclip, zhang2023adding} in image domain.
However, synthesizing data using iterative diffusion-based methods is computationally expensive and time-consuming.
Moreover, computational cost becomes even higher when it comes to video data.
Señorita-2M~\cite{zi2025senorita} recently released large-scale datasets for video editing, which are generated with such data pipelines.
However, it only includes a limited range (or variety) of editing and style transfer types.
Consequently, it remains insufficient for training the models to propagate a more diverse range of edits.
In contrast, our PropFly synthesizes diverse transformations from a limited set of real videos by employing on-the-fly data pair generation. 
Rather than relying on a fixed, precomputed paired dataset, PropFly generates structurally aligned yet semantically varied latent pairs on the fly, providing rich and flexible supervision for learning robust propagation-based video editing.

\section{Preliminary: Video Flow-Matching Models}
\label{sec:preliminary}

Our method is built upon Flow-Matching models \cite{lipman2022flow}. 
A neural network \(\theta\) (often consisting of DiT  blocks \cite{peebles2023scalable_dit}), conditioned on time \(t \sim U[0, 1]\) and text \(\mathbf{c}_\text{text}\), is trained to approximate the velocity vector field \(\mathbf{v}_t = \mathbf{x}_1 - \mathbf{x}_0\) that connects a data sample \(\mathbf{x}_0\) and a noise \(\mathbf{x}_1 \sim \mathcal{N}(\mathbf{0}, \mathbf{I})\).
It is trained by minimizing the flow-matching objective, a mean squared error between predicted and real velocities:
\begin{equation}
\label{eq:fm_objective}
\resizebox{0.9\linewidth}{!}{$
    \mathcal{L}_{\text{FM}} = \mathbb{E}_{t, (\mathbf{x}_0, \mathbf{c}_\text{text}), \mathbf{x}_1} \left[ \left\| (\mathbf{x}_1 - \mathbf{x}_0) - \mathbf{v}_\theta(\mathbf{x}_t, t, \mathbf{c}_\text{text}) \right\|^2 \right],
$}
\end{equation}
where \(\mathbf{x}_t = (1-t)\mathbf{x}_0 + t\mathbf{x}_1\).
During inference, a video is generated by solving the learned ordinary differential equation (ODE) backwards from \(t=1\) to \(t=0\) with a numerical solver. 
To control this process, Classifier-Free Guidance (CFG) \cite{ho2022classifier_cfg} is employed. 
A guided velocity \(\hat{\mathbf{v}}_\theta^\omega\) is computed at each step (\(t\)) using a guidance scale \(\omega\):
\begin{equation}
\label{eq:cfg_inference}
\begin{split}
\resizebox{0.9\linewidth}{!}{$    \hat{\mathbf{v}}_\theta^\omega = \mathbf{v}_\theta(\mathbf{x}_t, t, \emptyset) + \omega \cdot (\mathbf{v}_\theta(\mathbf{x}_t, t, \mathbf{c}_\text{text}) - \mathbf{v}_\theta(\mathbf{x}_t, t, \emptyset)),
$}
\end{split}
\end{equation}
where \(\emptyset\) is the null text token. 
This mechanism is crucial for enhancing text alignment and visual quality.

\section{Proposed Method}
\label{sec:proposed_method}

We propose a novel training pipeline for propagation-based video editing via on-the-fly supervision, PropFly.
As illustrated in Fig.~\ref{fig:main_overview}, our PropFly is designed to train an additional adapter, attached to the frozen VDM, to propagate the edits contained in the edited first frame to the entire source video.
Our PropFly pipeline consists of (a) Data Sampling \& Random Style Prompt Fusion, (b) On-the-fly Data Pair Generation, and (c) Guidance-Modulated Flow Matching.

\subsection{Model Architecture}

Our model consists of a frozen VDM backbone \(\theta\) with \(N_\text{B}\) DiT blocks trained via flow matching and an additional trainable adapter \(\phi\) (green blocks in Fig.~\ref{fig:main_overview}-(c)) with \(N_\text{B} / S_\text{in}\) DiT blocks, where \(S_\text{in}\) is the stride for condition injection.
To perform propagation-based video editing, the source video latent and the single edited first-frame latent are concatenated along the temporal dimension and fed as input to the adapter $\phi$, together with the text prompt \(\textbf{c}\). 
The adapter's output features are then injected into the frozen backbone $\theta$ at intervals of \(S_\text{in}\) to guide the generation.

\subsection{Data Sampling \& Random Style Prompt Fusion}
\label{subsec:random_style_prompt_fusion}
Our proposed training pipeline, PropFly, first randomly sample a pair of encoded video latent \(\mathbf{x}_0\) and its corresponding text caption \(\mathbf{c}_\text{text}\) from a video dataset (Fig.~\ref{fig:main_overview}-(a)).
To further enrich our on-the-fly training signals and expose the model to a wider variety of editing styles, we introduce Random Style Prompt Fusion (RSPF). 
By randomly fusing arbitrary style prompts \(\mathbf{c}_\text{style}\) (e.g., `\textit{in snow}' in Fig.~\ref{fig:onestep}) into the caption \(\mathbf{c}_\text{text}\) (e.g., `\textit{A bear walks}' in Fig.~\ref{fig:onestep}) of the original video, we can generate pairs with diverse combination of content and styles. 
The resulting augmented prompt \(\mathbf{c}_\text{aug} \coloneqq [\mathbf{c}_\text{style} | \mathbf{c_\text{text}}]\) is then used as a condition during our on-the-fly data pair generation and the training of adapter \(\phi\), ensuring robust training of propagation-based video editing with more diverse data pairs.

\subsection{On-the-fly Data Pair Generation}
\label{sec:on_the_fly_video_pair}

\paragraph{Key Observations.}
Classifier-Free Guidance (CFG) \cite{ho2022classifier_cfg} is a crucial component in the sampling process of diffusion models, primarily used to enhance visual quality and text alignment.
We extend the role of CFG beyond quality enhancement.
\textbf{Observation 1:} We observe that varying the CFG scales during the sampling of noised latents directly modulates the global visual properties of the output following the given text prompt, such as its artistic styles and color tones while preserving the overall context of videos (Fig.~\ref{fig:onestep}-(b)).
\textbf{Observation 2:} We observe that single-step estimations (Fig.~\ref{fig:onestep}-(c)) already give reasonable results. 
Our empirical results validate that supervision from a single-step clean latent estimation alone is sufficient to learn propagation-based video editing and that bypassing the full denoising process is possible.
These key observations suggest that a pre-trained VDM inherently understands how such global transformations are applied, and the amount of this transformation can be directly controlled with CFG, even from the single-step clean latent estimation.

\vspace{-5mm}
\paragraph{On-the-fly Data Pair Generation.}
Based on the above observations, we propose an on-the-fly data-pair generation pipeline (Fig.~\ref{fig:main_overview}-(b)) that leverages the varying scale of CFG for learning propagation-based global video editing.
With the sampled video \(x_0\) and augmented prompt \(\mathbf{c}_\text{aug}\), we add noise to \(\mathbf{x}_0\) at a random time \(t \sim U[0, 1]\) by linearly interpolating it with a noise vector \(\mathbf{x}_1 \sim \mathcal{N}(\mathbf{0}, \mathbf{I})\).
Since the pre-trained VDM backbone \(\theta\) is trained to predict the velocity vector \(\mathbf{v}_t = \mathbf{x}_1 - \mathbf{x}_0\) using the flow matching objective (Eq.~\ref{eq:fm_objective}), we can obtain a direct estimate of the clean latent \(\hat{\mathbf{x}}_{0|t}\) from any noised latent \(\mathbf{x}_t\) by reversing the path with the model's velocity prediction:
\begin{equation}
\label{eq:onestep}
    \hat{\mathbf{x}}_{0|t} = \mathbf{x}_t - t \cdot \mathbf{v}_\theta(\mathbf{x}_t, t, \mathbf{c_\text{aug}}).
\end{equation}
Here, we leverage the CFG scaling mechanism (Eq.~\ref{eq:cfg_inference}) that directly controls the intensity of the semantic edit for modulating \(\hat{\mathbf{x}}_{0|t}\). 
We then generate latent pair using two different scales, a low scale \(\omega_L\) (e.g., \(\omega_L=1.0\)) and a high scale \(\omega_H\) (e.g., \(\omega_H=7.0\)).
The source video latent \(\hat{\mathbf{x}}_{0|t}^{\text{low}}\) and the target (edited) video latent \(\hat{\mathbf{x}}_{0|t}^{\text{high}}\) are then generated as:
\begin{equation}
    \hat{\mathbf{x}}_{0|t}^{\text{low}} = \mathbf{x}_t - t \cdot \hat{\mathbf{v}}_\theta^\text{low}, \quad
    \hat{\mathbf{x}}_{0|t}^{\text{high}} = \mathbf{x}_t - t \cdot \hat{\mathbf{v}}_\theta^\text{high}, 
    \label{eq:target_latent}
\end{equation}
where \(\hat{\mathbf{v}}_\theta^\text{low}\) and \(\hat{\mathbf{v}}_\theta^\text{high}\) are CFG-scaled velocities (Eq.~\ref{eq:cfg_inference}) from the velocity predictions, \(\hat{\mathbf{v}}_\theta^\text{cond}=\mathbf{v}_\theta(\mathbf{x}_t, t, \mathbf{c_\text{aug}})\) and \(\hat{\mathbf{v}}_\theta^\text{uncond}=\mathbf{v}_\theta(\mathbf{x}_t, t, \emptyset)\), by using \(\omega_L\) and \(\omega_H\), respectively.

\begin{figure*}[t!]
    \includegraphics[width=1\linewidth] {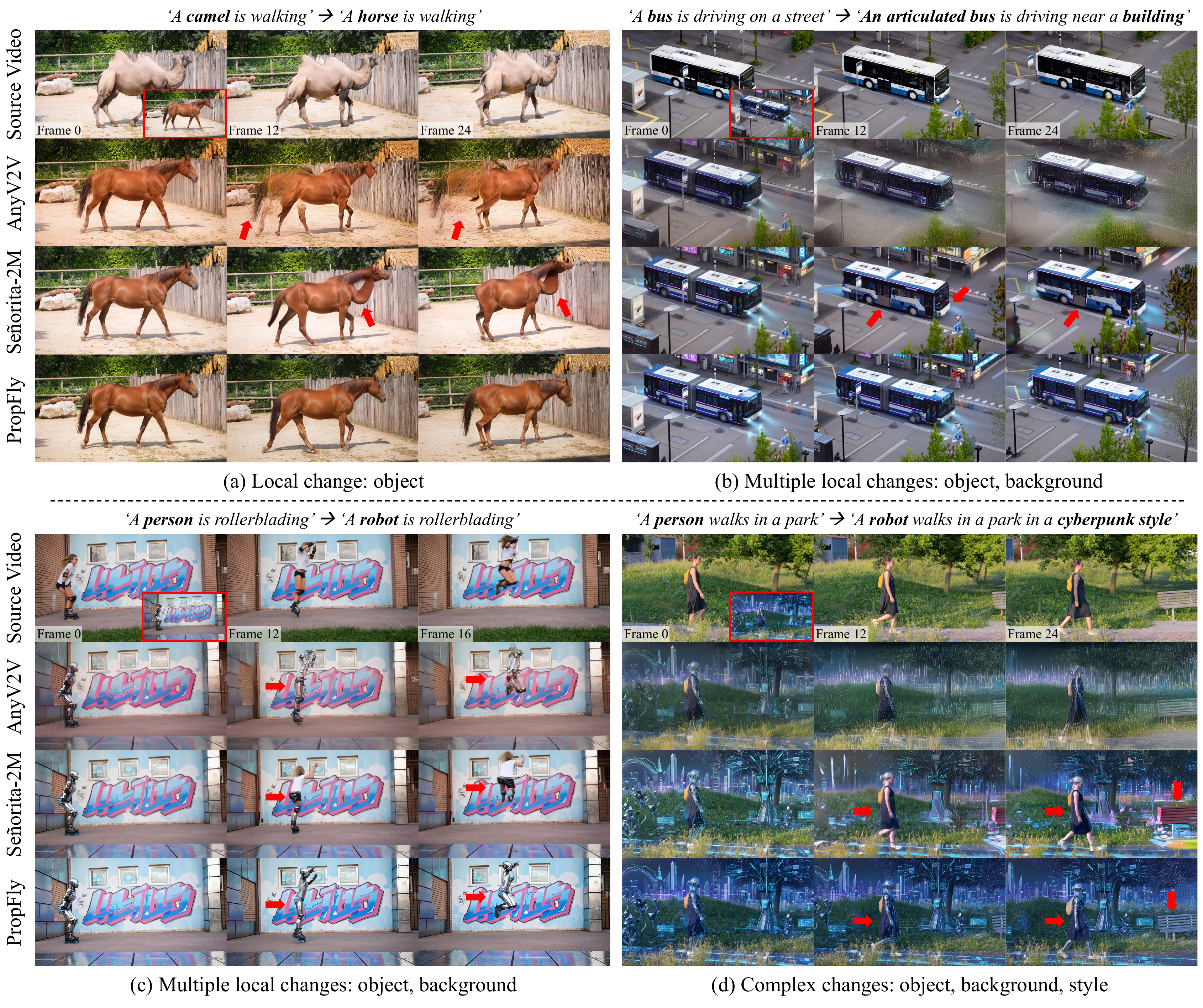}
    \caption{Qualitative comparison against propagation-based baselines AnyV2V \cite{ku2024anyv2v} and Señorita-2M~\cite{zi2025senorita}. Our PropFly successfully propagates diverse edits (including object, background, and style changes) while preserving the motion of the source videos. In contrast, the baseline methods often fail to propagate the edits accurately or introduce severe visual artifacts. Zoom in for better visualization.}
    \label{fig:qualitative_davis}
\end{figure*}

This one-step estimation strategy with CFG scaling ensures that the source latent \(\hat{\mathbf{x}}_{0|t}^{\text{low}}\) and target latent \(\hat{\mathbf{x}}_{0|t}^{\text{high}}\) are semantically different, but well-aligned in their structure and motion, as they originate from the same velocity prediction. 
The crucial element is not the visual fidelity of the one-step latents, but the \textit{semantic difference} between them, which provides a clean signal for guiding the propagation. 
By learning this generalized transformation, our model achieves strong generalization and is able to perform a wide range of edits, from local to complex, as shown in Fig.~\ref{fig:qualitative_davis}.
Also, this process only adds a modest computational overhead compared to generating edited videos via full sampling, yet it overcomes the dataset scarcity problem by enabling infinite generation of diverse training pairs through randomly sampled \(\mathbf{x}_1\) and \(t\).
As detailed in Sec.~\ref{sec:distillation_loss}, the source latent $\hat{\mathbf{x}}_{0|t}^{\text{low}}$ is used as the structural condition, while the target latent $\hat{\mathbf{x}}_{0|t}^{\text{high}}$ provides both the style condition (its first frame) and the supervision target (its velocity $\hat{\mathbf{v}}_\theta^\text{high}$).

\begin{algorithm}
\small
\caption{PropFly Training Pipeline}
\label{alg:training}
\begin{algorithmic}[1] 
\Require Frozen VDM $\theta$, Trainable adapter $\phi$
\Require VAE Encoder $\mathcal{E}$
\Require Training dataset $\mathcal{D}$, Set of style prompts $\mathcal{A}_\text{style}$
\Require Low/High CFG scales $\omega_L, \omega_H$
\Require Learning rate $\eta$, Number of training iterations $N$
\For{$i=1$ \textbf{to} $N$}
    \vspace{1mm}
    \State \textbf{\textit{1. Data Preparation \& RSPF}}
    \State $(\mathbf{x}_\text{data}, \mathbf{c_\text{text}}) \sim \mathcal{D}$ \Comment{Sample a video-text pair}
    \State $\mathbf{c}_\text{style} \sim \mathcal{A}_\text{style}$ \Comment{Sample a random style prompt}
    \State $\mathbf{c}_\text{aug} \leftarrow [\mathbf{c}_\text{style} | \mathbf{c_\text{text}}]$ \Comment{RSPF in Sec.~\ref{subsec:random_style_prompt_fusion}}
    \State $\mathbf{x}_0 \leftarrow \mathcal{E}(\mathbf{x}_\text{data})$ \Comment{Encode video to latent space}
    \State $t \sim U[0, 1]$, $\mathbf{x}_1 \sim \mathcal{N}(0, \mathbf{I})$ \Comment{Sample time \& noise}
    \State $\mathbf{x}_t \leftarrow (1-t)\mathbf{x}_0 + t\mathbf{x}_1$ \Comment{Add noise}
    \vspace{1mm}
    \State \textbf{\textit{2. On-the-fly Data Pair Generation}}
    \State $\hat{\mathbf{v}}_\theta^\text{uncond}, \; \hat{\mathbf{v}}_\theta^\text{cond} \leftarrow \mathbf{v}_\theta(\mathbf{x}_t, t, \emptyset), \; \mathbf{v}_\theta(\mathbf{x}_t, t, \mathbf{c_\text{aug}})$ 
    \State \Comment{pre-trained VDM prediction}
    \State $\hat{\mathbf{v}}_\theta^\text{low} \leftarrow \hat{\mathbf{v}}_\theta^\text{uncond} + \omega_L \cdot (\hat{\mathbf{v}}_\theta^\text{cond} - \hat{\mathbf{v}}_\theta^\text{uncond})$ 
    \State $\hat{\mathbf{v}}_\theta^\text{high} \leftarrow  \hat{\mathbf{v}}_\theta^\text{uncond} + \omega_H \cdot (\hat{\mathbf{v}}_\theta^\text{cond} - \hat{\mathbf{v}}_\theta^\text{uncond})$ 
    \State $\hat{\mathbf{x}}_{0|t}^{\text{low}} \leftarrow \mathbf{x}_t - t \cdot \hat{\mathbf{v}}_\theta^\text{low}$ \Comment{"Source" latent (Eq. \ref{eq:target_latent})}
    \State $\hat{\mathbf{x}}_{0|t}^{\text{high}} \leftarrow \mathbf{x}_t - t \cdot \hat{\mathbf{v}}_\theta^\text{high}$ \Comment{"Target" latent (Eq. \ref{eq:target_latent})}
    \vspace{1mm}
    \State \textbf{\textit{3. Guidance-Modulated Flow Matching}}
    \State $\hat{\mathbf{v}}_{\theta,\phi} \leftarrow \mathbf{v}_{\theta, \phi}(\mathbf{x}_t, t, \mathbf{c_\text{aug}}, \hat{\mathbf{x}}_{0|t}^{\text{low}}, \hat{\mathbf{x}}_{0|t}^{\text{high}}[0])$ 
    \State $\mathcal{L}_{\text{GMFM}} \leftarrow \left\| \hat{\mathbf{v}}_{\theta,\phi} - \text{sg}\{\hat{\mathbf{v}}_\theta^\text{high}\} \right\|^2$ \Comment{GMFM loss (Eq. \ref{eq:distill_loss})}
    \State $\phi \leftarrow \phi - \eta \cdot \nabla_\phi \mathcal{L}_{\text{GMFM}}$ \Comment{Update adapter parameters}
\EndFor
\end{algorithmic}
\end{algorithm}

\subsection{Guidance-Modulated Flow Matching}
\label{sec:distillation_loss}

For training propagation-based video editing with the on-the-fly generated data pairs, we introduce a Guidance-Modulated Flow Matching (GMFM) loss (Fig.~\ref{fig:main_overview}-(c)).
Our model predicts the velocity conditioned on: 
(i) the \textit{entire} source video \(\hat{\mathbf{x}}_{0|t}^{\text{low}}\) (as structural guidance), (ii) the \textit{first frame} of the target video \(\hat{\mathbf{x}}_{0|t}^{\text{high}}[0]\) (as the visual style guidance), and (iii) the style fused text prompt \(\mathbf{c}_\text{aug}\).
The full velocity prediction by our model, $\hat{\mathbf{v}}_{\theta, \phi}$, is thus formulated as:
\begin{equation}
    \hat{\mathbf{v}}_{\theta, \phi} = {\mathbf{v}}_{\theta, \phi}(\mathbf{x}_t, t, \mathbf{c}_\text{aug}, \hat{\mathbf{x}}_{0|t}^{\text{low}}, \hat{\mathbf{x}}_{0|t}^{\text{high}}[0]).
\label{eq:velocity_pred}
\end{equation}
As indicated in Eq.~\ref{eq:velocity_pred}, we feed the same noised latent \(\mathbf{x}_t\), used in the generation of the on-the-fly data pairs, rather than sampling a new noise and timestep. 
By feeding \(\mathbf{x}_t\) to our model, VDM backbone \(\theta\) can easily reconstruct its original prediction \(\hat{\mathbf{v}}_\theta^{\text{cond}}=\mathbf{v}_\theta(\mathbf{x}_t, t, \mathbf{c_\text{aug}})\), and thus the adapter \(\phi\) can concentrate exclusively on learning to propagate the transformation of \(\hat{\mathbf{x}}_{0|t}^{\text{low}}\) into \(\hat{\mathbf{x}}_{0|t}^{\text{high}}\).
Then, our model is trained to match the VDM's high-CFG velocity vectors \(\hat{\mathbf{v}}_\theta^\text{high}\), which encapsulate the semantic transformation introduced by varying CFG scales. 
This forms our GMFM loss, \(\mathcal{L}_{\text{GMFM}}\):
\begin{equation}
    \mathcal{L}_{\text{GMFM}} = \mathbb{E}_{t, (\mathbf{x}_0, \mathbf{c}_\text{text}), \mathbf{x}_1, \mathbf{c}_\text{style}}\left[ \left\| \hat{\mathbf{v}}_{\theta, \phi} - \text{sg}\{\hat{\mathbf{v}}_\theta^\text{high}\}\right\|^2 \right],
\label{eq:distill_loss}
\end{equation}
where \(\text{sg}\{\cdot\}\) denotes a stop-gradient operation, as the VDM backbone is frozen.
This strategy effectively guides the adapter to associate the visual style of the \textit{first frame} (\(\hat{\mathbf{x}}_{0|t}^{\text{high}}[0]\)) with the complete semantic transformation that the pre-trained VDM already knows how to perform.
The overall training pipeline of PropFly is described in Alg.~\ref{alg:training}.

\begin{table}
    \scriptsize
    \centering
    \caption{Quantitative comparison on the EditVerseBench-Appearance subset \cite{ju2025editverse}. `\textbf{Te}' and `\textbf{Pr}' in the second column denote \textbf{text-guided} and \textbf{propagation-based} video editing (VE) methods, respectively. We evaluate video quality (Pick), text alignment (Frame \& Video), and temporal consistency (CLIP \& DINO). $\uparrow$ indicates higher is better. 
    }
    \setlength{\tabcolsep}{3pt}  
\begin{tabular}{lccccccc}
\toprule
     Methods & VE & Pick {$\uparrow$} & Frame {$\uparrow$} & Video {$\uparrow$} & CLIP {$\uparrow$} & DINO {$\uparrow$} & {\#} Param.\\
     \midrule
    TokenFlow \cite{tokenflow2023} & Te & 20.02 & 26.80 & 24.30 & 98.68 & 98.78 & 0.9B\\
    STDF \cite{yatim2024space_stdf} & Te & 19.73 & 26.35 & 23.64 & 96.37 & 96.00 & 1.7B\\
    InsV2V \cite{cheng2023consistent_insv2v} & Te & 19.55 & 26.14 & 23.33 & 97.15 & 96.67 & 0.9B \\
    Lucy Edit \cite{wan2025wan_lucy} & Te & 19.52 & 26.64 & 23.74 & 98.51 & 98.43 & - \\
    VACE \cite{vace} & Te & 20.04 & 27.33 & 24.50 & 98.96 & 98.79 & 14B \\
    EditVerse \cite{ju2025editverse} & Te & 20.06 & 27.95 & 25.48 & 98.58 & 98.56 & - \\
    Runway \cite{RunwayAleph2025} & Te & 20.19 & 28.18 & 24.96 & 98.82 & 98.39 & - \\ 
    \midrule
    CCEdit \cite{feng2024ccedit} & Pr & 19.59 & 27.20 & 25.32 & 96.55 & 95.08 & 1.6B \\
    AnyV2V \cite{ku2024anyv2v} & Pr & 19.78 & 28.19 & 25.34 & 95.97 & 97.73 & 1.3B  \\
    Señorita \cite{zi2025senorita} & Pr & 19.69 & 27.36 & 24.53 & 98.04 & 98.03 & 5B  \\
    \midrule
    PropFly-1.3B & Pr & 20.35 & 28.37 & 25.37 & 99.03 & 98.83 & 1.3B  \\
    \textbf{PropFly-14B} & Pr & \textbf{20.42} & \textbf{28.71} & \textbf{26.05} & \textbf{99.21} & \textbf{99.05} & 14B  \\
    \bottomrule
\end{tabular}

    \label{tab:editverse}
\end{table}

\begin{table}
    \scriptsize
    \centering
    \caption{Quantitative comparison on the TGVE benchmark \cite{wu2023tgvecvpr}. `\textbf{Te}' and `\textbf{Pr}' in the second column denote \textbf{text-guided} and \textbf{propagation-based} video editing (VE) methods, respectively. We evaluate video quality (Pick), temporal consistency (CLIP), and text alignment ($\text{ViCLIP}_{dir}$ \& $\text{ViCLIP}_{out}$). 
    }
    \setlength{\tabcolsep}{5pt}  
\begin{tabular}{lcccccc}
\toprule
    Methods & VE & Pick {$\uparrow$} & CLIP {$\uparrow$} & $\text{ViCLIP}_{dir}$  {$\uparrow$} & $\text{ViCLIP}_{out}$ {$\uparrow$} \\
     \midrule
    TAV \cite{wu2023tune} & Te & 20.36 & 0.924 & 0.162 & 0.243 \\
    SDEdit \cite{meng2021sdedit} & Te & 20.18 & 0.896 & 0.172 & 0.253 \\
    STDF \cite{yatim2024space_stdf} & Te & 20.40 & 0.933 & 0.110 & 0.226 \\
    Fairy \cite{wu2024fairy} & Te & 19.80 & 0.933 & 0.164 & 0.208 \\
    InsV2V \cite{cheng2023consistent_insv2v} & Te & 20.76 & 0.911 & 0.208 & 0.262 \\
    EVE \cite{singer2024video_eve} & Te & 20.76 & 0.922 & 0.221 & 0.262 \\
    \midrule
    CCEdit \cite{feng2024ccedit} & Pr & 20.46 & 0.917 & 0.218 & 0.260 \\
    AnyV2V \cite{ku2024anyv2v} & Pr & 20.55 & 0.933 & 0.208 & 0.246 \\
    Señorita \cite{zi2025senorita} & Pr & 20.54 & 0.961 & 0.220 & 0.254 \\
    \midrule
    PropFly-1.3B & Pr & 20.72 & 0.965 & 0.221 & 0.269 \\
    \textbf{PropFly-14B} & Pr & \textbf{21.19} & \textbf{0.978} & \textbf{0.228} & \textbf{0.278} \\
    \bottomrule
\end{tabular}
    \label{tab:tgve}
\end{table}

\section{Experiments}
\label{sec:experiments}

\subsection{Implementation Details}
\label{sec:implementation_details}

We use the frozen Wan2.1~\cite{wan2025wan_lucy} T2V model as the backbone and attach a trainable VACE adapter~\cite{vace}, initialized from the VACE weights trained for I2V generation.
For our PropFly-14B (initialized from Wan2.1-14B), the number of blocks \(N_\text{B}\) is \(35\) and the adapter injection stride \(S_\text{in}\) is \(5\), and PropFly-1.3B (initialized from Wan2.1-1.3B) has \(N_\text{B}=30\) and \(S_\text{in}=2\).
We train our models using a combined dataset of videos from Youtube-VOS \cite{xu2018youtube} and manually collected 3,000 videos from Pexels~\cite{pexels_website}, with captions generated by Qwen2.5-VL~\cite{bai2025qwen2}. 
Our PropFly is trained for 50K iterations at a resolution of $480 \times 832$  using the AdamW optimizer~\cite{loshchilov2017decoupled} with a learning rate of $1 \times 10^{-5}$ and a global batch size of 48. 
For our on-the-fly data pair generation, we use the CFG scales of $\omega_H = 7$ and $\omega_L = 1$.
During inference, we feed the condion features (the edited first frame concatenated with the entire source video along the temporal axis) to our adapter.
We then perform denoising using the UniPC scheduler~\cite{zhao2023unipc} with 25 steps, which takes approximately 120 seconds for PropFly-14B and 30 seconds for PropFly-1.3B.
We utilize the Gemini 2.5 Flash Image model \cite{comanici2025gemini} to synthesize the edited frames for propagation in our experiments, unless explicitly provided in the benchmark dataset
Our experiments are conducted on 4 NVIDIA A100 80GB GPUs.
More details are described in \textit{Suppl.}


\subsection{Comparison to Other Methods}

\paragraph{Qualitative Comparison.}
We qualitatively compare PropFly against other SOTA propagation-based video editing methods, AnyV2V \cite{ku2024anyv2v} and Señorita-2M~\cite{zi2025senorita}.
In Fig.~\ref{fig:qualitative_davis}, we compare diverse editing scenarios, ranging from local object change to complex, multiple edits on the object, background, and style.
AnyV2V \cite{ku2024anyv2v}, which is a zero-shot method, introduces significant visual artifacts and fails to propagate edits onto moving objects.
For example, it ruins the structure of the horse in (a), and fine details in the later frames are blurred and destroyed in (b) and (d).
Señorita-2M~\cite{zi2025senorita}, trained on a large-scale paired dataset, struggles with complex edits and temporal consistency.
For example, Señorita-2M fails to maintain the person-to-robot transformation in (c) and (d), and the original structure of the bench in (d). 
In contrast, our PropFly robustly propagates various types of edits while maintaining the main object's motion and the context of the background.
For instance, PropFly successfully propagates the transformation of the camel into a horse in (b) and the person into a robot in (d), all while perfectly preserving their complex original motions.
Our method also correctly handles occlusions, propagating the style to later-unoccluded regions like the bench in (d).
\textbf{More results are provided in \textit{Suppl.}}

\vspace{-5mm}
\paragraph{Quantitative Comparison.}
We conduct quantitative evaluations of our method by comparing with several SOTA baselines, which are grouped into two categories: (i) text-guided methods \cite{tokenflow2023, yatim2024space_stdf, cheng2023consistent_insv2v, wan2025wan_lucy, ju2025editverse, RunwayAleph2025}, and (ii) propagation-based methods \cite{zi2025senorita, feng2024ccedit, ku2024anyv2v}.
We evaluate the video editing methods on the EditVerseBench-Appearance subset. 
This subset is derived from the full EditVerseBench \cite{ju2025editverse}, which is a recent evaluation benchmark for instruction-based V2V editing, by selecting 11 tasks relevant to visual appearance editing (e.g., stylization, background, object modification), while excluding the tasks that are not relevant to the scope of this work (e.g., camera view change, depth-to-video).
We evaluate all methods using a suite of standard metrics: (i) video quality is assessed using frame-wise Pick \cite{kirstain2023pick}, (ii) text alignment is measured using both CLIP \cite{radford2021learning_clip} (frame-level) and ViCLIP \cite{wang2023internvid_viclip} (video-level), and (iii) temporal consistency is evaluated in terms of frame-to-frame similarity in both CLIP \cite{radford2021learning_clip} and DINO \cite{caron2021emerging_dino} feature spaces.
As shown in Table~\ref{tab:editverse}, our PropFly-14B achieves SOTA performance across all five metrics. Our method surpasses strong baselines, including text-based methods (EditVerse \cite{ju2025editverse} and Runway Aleph \cite{RunwayAleph2025}) and propagation-based methods (Señorita-2M \cite{zi2025senorita} and AnyV2V \cite{ku2024anyv2v}).
Also, our PropFly-1.3B outperforms baselines on most metrics, validating the effectiveness of our training pipeline.

We also evaluate our method on the TGVE benchmark \cite{wu2023tgvecvpr} on a set of video editing tasks, including `style', `object', `background', and `multiple' changes. 
We follow the evaluation protocol from TGVE \cite{wu2023tgvecvpr}, assessing three criteria: (i) video quality using frame-wise Pick \cite{kirstain2023pick}, (ii) temporal consistency via frame-to-frame CLIP \cite{radford2021learning_clip} feature similarity and (iii) text alignment using both Text-Video Direction Change Similarity (denoted as $\text{ViCLIP}_{dir}$) and Output Text-Video Direction Similarity (denoted as $\text{ViCLIP}_{out}$), as measured by ViCLIP \cite{wang2023internvid_viclip}.
As shown in Table~\ref{tab:tgve}, our PropFly \textit{significantly} outperforms other methods across all reported metrics on the TGVE benchmark \cite{wu2023tgvecvpr}.




\begin{table}[t!]
    \footnotesize
    \centering
    \caption{Ablation study of our key components on the EditVerseBench-Appearance subset \cite{ju2025editverse}.}
    \setlength{\tabcolsep}{1pt}  
\begin{tabular}{lccccc}
\toprule
    
    Method & Pick {\small$\uparrow$} & Frame {\small$\uparrow$} & Video {\small$\uparrow$} & CLIP {\small$\uparrow$} & DINO {\small$\uparrow$} \\
     \midrule
    w/ Full sampling & 19.75 & 27.20 & 24.77 & 98.77 & 98.51 \\
    w/ FM loss (Eq.~\ref{eq:fm_objective}) & 19.50 & 26.33 & 21.98 & 98.52 & 98.29 \\
    w/o RSPF (Sec.~\ref{subsec:random_style_prompt_fusion}) & 20.28 & 28.35 & \textbf{25.61} & 98.96 & 98.55 \\
    w/ Paired dataset & 19.53 & 27.12 & 24.69 & 98.13 & 97.85 \\
    \midrule
    \textbf{PropFly-1.3B} & \textbf{20.35} & \textbf{28.37} & 25.37 & \textbf{99.03} & \textbf{98.63} \\
    \bottomrule
\end{tabular}
    \label{tab:ablation_editverse}
\end{table}

\begin{figure}[t!]
    \centering
    \includegraphics[width=\linewidth] {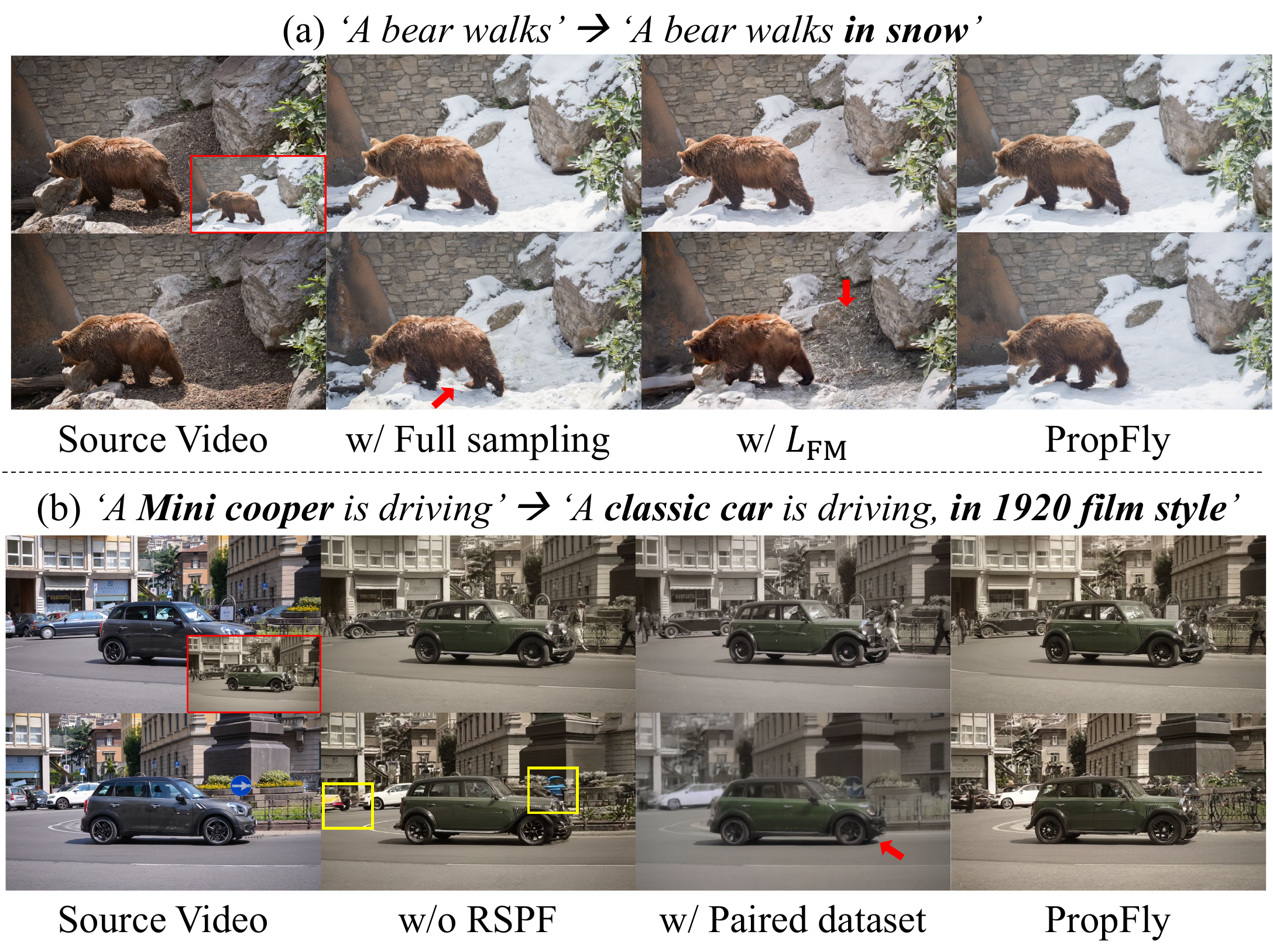}
    \caption{Visual results showing the effect of our key components. (a) Baseline trained with full sampling fails to align object motion, while the baseline trained with the conventional FM objective fails to propagate the edit. (b) Baselines trained without our RSPF or with the paired dataset lack generalization, failing to perform complex edits. In contrast, our PropFly achieves robust propagation performance and high-fidelity edits. Zoom-in for details.}
    \label{fig:ablation}
    \vspace{-3mm}
\end{figure}

\subsection{Ablation Study}

We conduct ablation studies to validate the key components of PropFly on EditVerseBench-Appearance \cite{ju2025editverse} subset. We utilize the Wan2.1-1.3B model \cite{wan2025wan_lucy} as our backbone.

\vspace{-4mm}
\paragraph{One-step Clean Latent Estimation vs. Full Sampling.}
We validate our one-step estimation against a baseline using full sampling (i.e., an iterative ODE solve from \(t\) to 0 for each CFG scale) to generate the source and edited pairs.
In Table~\ref{tab:ablation_editverse}, the full sampling baseline demonstrates inferior performance, producing videos with severe motion misalignment (e.g., the bear is not moving) in Fig.~\ref{fig:ablation}. 
This is because the two independent, iterative sampling paths (low-CFG and high-CFG) accumulate numerical errors and diverge, often resulting in unaligned pairs. 
In contrast, our one-step estimation is a direct calculation from the identical latent $\mathbf{x}_t$, ensuring the source and target are perfectly aligned and providing a clean supervision signal.

\vspace{-5mm}
\paragraph{GMFM vs. Standard FM.}
We validate our GMFM loss (Eq.~\ref{eq:distill_loss}) over the standard flow-matching (FM) objective (Eq.~\ref{eq:fm_objective}) of the baseline. 
As shown in Table~\ref{tab:ablation_editverse} and Fig.~\ref{fig:ablation}, the baseline trained with the regular FM loss fails to propagate the edited part in the first frame (the snow from the first frame disappears in the later frames) since the FM loss trains the model to reconstruct the original video, creating a contradictory objective.
In contrast, our GMFM loss trains the adapter to reconstruct the target transformation derived from our on-the-fly pairs, providing the correct supervisory signal and leading to successful edit propagation.

\vspace{-5mm}
\paragraph{Random Style Prompt Fusion (RSPF).}
We validate our RSPF by training a baseline without it. 
As shown in Table~\ref{tab:ablation_editverse}, this baseline shows a clear performance degradation in the video quality metric (Pick Score) and fails to align with the reference style.
For example, in Fig.~\ref{fig:ablation}, it fails to consistently apply a `\textit{1920s film style}', allowing colorful cars (in yellow boxes) to appear in later frames, which breaks the monochrome aesthetic.
This confirms that our RSPF provides rich content-style combinations for learning complex transformations and significantly improves generalization to unseen edits at inference time.

\vspace{-5mm}
\paragraph{PropFly vs. Paired Dataset.}
To validate the quality of our on-the-fly supervision, we compare PropFly against the baseline trained on a paired video editing dataset, Señorita-2M \cite{zi2025senorita}.
In Table~\ref{tab:ablation_editverse}, our PropFly significantly outperforms the baseline trained with ground-truth paired data,
Also in Fig.~\ref{fig:ablation}, the supervised baseline fails to maintain the `\textit{Mini cooper to classic car}' transformation in later frames.
This result confirms that our on-the-fly supervision from pre-trained VDMs provides more diverse editing cases, leading to robust training of propagation-based video editing. 

\section{Conclusion}
\label{sec:conclusion}

In this paper, we introduced PropFly, a novel training pipeline for propagation-based global video editing that circumvents the need for precomputed paired training data. 
Our method leverages a pre-trained, frozen video flow-matching model to generate source and edited video pairs on the fly. 
We create these pairs by exploiting a key property of Classifier-Free Guidance: varying the CFG scale produces two aligned video latents that share the same motion and structure but have a distinct semantic gap, where the low CFG result can be used as the source and the high CFG result can be used as the target.
A trainable adapter is then trained using our proposed guidance-modulated flow matching (GMFM) loss. 
With this loss, the adapter effectively learns to replicate the pre-trained model's text-guided transformations using only visual conditions, specifically the full source video for structure and the single edited frame for style.
Extensive experiments and ablation studies validate that PropFly, trained without any paired video data, significantly outperforms other video editing baselines. 
We believe our framework presents a promising new paradigm for training powerful and generalizable video editing models by alleviating the need for large-scale paired data.
\section{Acknowledgement}
This work was supported by Institute of Information \& communications Technology Planning \& Evaluation (IITP) grant funded by the Korean Government [Ministry of Science and ICT (Information and Communications Technology)] (Project Number: RS-2022-00144444, Project Title: Deep Learning Based Visual Representational Learning and Rendering of Static and Dynamic Scenes, 100\%).


{
    \small
    \bibliographystyle{ieeenat_fullname}
    \bibliography{main}
}

\clearpage
\setcounter{page}{1}
\maketitlesupplementary
\appendix


\section{Introduction}

In this supplementary material, we provide additional details omitted from the main paper.
Sec.~\ref{sec:supp_implementation_details} and Sec.~\ref{sec:supp_evaluation_details} elaborate on the implementation details and evaluation protocols, respectively.
Sec.~\ref{sec:supp_ablation_study} presents in-depth ablation studies, including analyses on CFG scaling and data pair quality, followed by a discussion of limitations in Sec.~\ref{sec:supp_limitations}.
Finally, Sec.~\ref{sec:supp_qualitative_comparison} provides extensive qualitative comparisons on the DAVIS~\cite{Perazzi2016} and EditVerseBench~\cite{ju2025editverse} datasets to further demonstrate the capabilities of PropFly.
We strongly recommend viewing the accompanying \href{./Supplementary_Videos/comparison_videos.html}{propagation\_comparison.html} and \href{./Supplementary_Videos/PropFly_videos.html}{PropFly\_videos.html} files located in the \textit{\textbf{Supplementary\_Videos}} directory to fully assess the temporal consistency and visual quality of our results.

\section{Implementation Details}
\label{sec:supp_implementation_details}

\subsection{Training Details}

We train our PropFly models for a total of 50,000 iterations at a fixed resolution of $480 \times 832$ with 33 frames. 
Optimization is performed using the AdamW optimizer~\cite{loshchilov2017decoupled} with $\beta_1 = 0.9$, $\beta_2 = 0.999$, and a weight decay of $0.1$. 
The learning rate is held constant at $1 \times 10^{-5}$.
The training is conducted on 4 NVIDIA A100 (80GB) GPUs. 
For PropFly-14B, the training process takes approximately 12 days with a global batch size of 4. 
For PropFly-1.3B, it requires approximately 2.5 days with a global batch size of 48.
We utilize Bfloat16 precision for both PropFly-14B and PropFly-1.3B to optimize memory usage and training speed.

The pre-trained Wan2.1 backbone~\cite{wan2025wan_lucy} remains frozen. 
We only finetune the following parameters within the DiT adapter blocks: patch embedding parameters, linear projection layers within the attention blocks, including \texttt{to\_q}, \texttt{to\_k}, \texttt{to\_v}, and \texttt{to\_out}.

\subsection{Inference Details}

For all inference results, we use a single Classifier-Free Guidance (CFG) scale of $\omega = 1.0$. 
All source videos were pre-processed by resizing and center-cropping them to a $480 \times 832$ resolution.
We perform denoising using the UniPC scheduler~\cite{zhao2023unipc} with 25 sampling steps.

\subsection{Computational Complexity}

The primary overhead of our PropFly training pipeline stems from the added sampling process required for on-the-fly data pair generation.
Since this sampling is fully decoupled from the loss calculation, no gradient backpropagation is required for this step. 
Consequently, the necessary GPU memory footprint remains unchanged compared to the baseline VDM training setup.
The required training time of one iteration for our PropFly-1.3B is 4.95 seconds, compared to the training time with paired datasets of 4.71 seconds.
This represents an overhead of approximately 5.1\% per iteration.
Note that training with paired datasets requires encoding both source and edited videos, while PropFly only needs to encode the original video, thereby reducing the training time gap.
As PropFly is a training pipeline designed only to train a lightweight adapter, the computational complexity for inference remains the same as the underlying VDM architecture with the adapter.

\subsection{Style Prompt Set}

\begin{table*}[t!]
    \footnotesize
    \centering
    \caption{Style prompt set used for RSPF during the training of PropFly.}
    \begin{tabular}{lllll}
\toprule
      \textbf{Weather} & \textbf{Artistic Style} & \textbf{Material} & \textbf{Mood} & \textbf{Backgrounds} \\
\midrule
        snowy &  in the style of Van Gogh &   made of glass &        monochrome &                   Gradient Background \\
        rainy &     in the style of Monet & made of crystal &    vibrant colors &            Moving Particle Background \\
    drizzling &   in the style of Picasso &     made of ice &     pastel colors &          Geometric Pattern Background \\
        foggy &      in the style of Dali &         glowing &      muted colors &                  Neon Line Background \\
        misty & in the style of Rembrandt &     holographic &            dreamy &       Abstract Digital Art Background \\
        sunny &   in the style of Hokusai &      iridescent &          ethereal &                Matrix Code Background \\
     overcast &   in the style of Ukiyo-e &        metallic &             eerie &                     Plexus Background \\
       stormy &             Impressionism &          chrome &            serene &          Watercolor Splash Background \\
 thunderstorm &                Surrealism &            gold &         nostalgic &        Smoke or Fog Effect Background \\
partly cloudy &                    Cubism &          silver & dramatic lighting &               Bokeh Effect Background \\
        windy &                   Pop Art &          bronze &     soft lighting &                 Cloudy Sky Time-lapse \\
         hazy &               Art Nouveau &          wooden &         cinematic &                      Starry Night Sky \\
  golden hour &       watercolor painting &          marble &              epic &                      Calm Ocean Waves \\
    blue hour &              oil painting &           stone &        minimalist &                 Rainy Day Window View \\
        night &          acrylic painting &          fabric &        maximalist &                        Snowfall Scene \\
         dawn &                    sketch &          velvet &    dark and moody &                Sunbeams in the Forest \\
         dusk &          charcoal drawing &            clay &    light and airy &                     Aerial Drone View \\
       sunset &       pen and ink drawing &         origami &                   &                        Sunset Scenery \\
      sunrise &               digital art &                 &                   &                   Tranquil Lake Scene \\
   midday sun &               concept art &                 &                   &                Four Seasons Landscape \\
       aurora &                 pixel art &                 &                   &              Paper Texture Background \\
              &                vector art &                 &                   &            Old Film Effect Background \\
              &                  low poly &                 &                   &                 Wood Grain Background \\
              &                           &                 &                   &             Fabric Texture Background \\
              &                           &                 &                   &              Concrete Wall Background \\
              &                           &                 &                   &       Water Drops on Glass Background \\
              &                           &                 &                   & Glitter and Sparkle Effect Background \\
              &                           &                 &                   &                 City Night Time-lapse \\
              &                           &                 &                   &           Library or Study Background \\
              &                           &                 &                   &                  Cozy Cafe Background \\
              &                           &                 &                   &    High-Tech Circuit Board Background \\
              &                           &                 &                   &             Newsroom Style Background \\
              &                           &                 &                   &  Stage Light and Spotlight Background \\
              &                           &                 &                   &   Clean White/Black Studio Background \\
\bottomrule
\end{tabular}
    \label{tab:style_prompt}
\end{table*}

To implement the Random Style Prompt Fusion (RSPF) introduced in Sec.~\ref{subsec:random_style_prompt_fusion}, we curated a diverse collection of style descriptors to enrich our on-the-fly supervision.
We generated an initial candidate list using the Gemini 2.5 Flash model~\cite{comanici2025gemini} and manually filtered the results to ensure distinctiveness and visual impact.
The final set comprises 113 phrases spanning a wide range of categories, including weather, artistic styles, materials, mood, and backgrounds. 
The complete list is provided in Table~\ref{tab:style_prompt}.

\section{Evaluation Details}
\label{sec:supp_evaluation_details}

\subsection{Edited First Frame Generation}

For edited first frame synthesis, we provide the Gemini 2.5 Flash Image model \cite{comanici2025gemini} with the original first frame and the target text prompt.
For fair comparison of our PropFly against other propagation-based video editing methods, AnyV2V \cite{ku2024anyv2v} and Señorita-2M \cite{zi2025senorita}, we use the same edited first frames.
Since CCEdit \cite{feng2024ccedit} propagates the edits from the center frame, we utilized the edited center frame from the EditVerse results.
Additionally, for the `Propagation' category within EditVerseBench, we employ the officially provided edited first frames to align with standard evaluation protocols.

\subsection{EditVerseBench}

For quantitative comparison, we evaluate our PropFly on the EditVerseBench \cite{ju2025editverse}, a recent benchmark for instruction-guided video editing.
The full benchmark comprises 100 videos (50 horizontal and 50 vertical) spanning 20 distinct editing categories, resulting in 200 total video-instruction pairs.
As our method focuses on visual appearance and style propagation, we utilize a subset of the full benchmark relevant to our setting, referred to as EditVerseBench-Appearance in the main paper. We selected 11 categories relevant to our scope and excluded tasks unrelated to appearance modification (e.g., `Change camera pose', `Detection', `Pose-to-video', `Depth-to-video', `Edit with mask').
The 11 selected categories are: `Add object', `Remove object', `Change object', `Stylization', `Propagation', `Change background', `Change color', `Change material', `Add effect', `Change weather', `Combined tasks'.

For assessing video quality, we calculate the PickScore~\cite{kirstain2023pick} using the CLIP ViT-H/14~\cite{radford2021learning_clip} backbone. 
For text-frame alignment, we average the cosine similarities calculated across all frames and the text instruction, encoded using the CLIP ViT-L/14~\cite{radford2021learning_clip} backbone. 
For text-video alignment, we calculate the cosine similarity between the video and text prompt, encoded using the ViCLIP-InternVid-10M-Flt~\cite{wang2023internvid_viclip} checkpoint. 
For temporal consistency, we calculate the average cosine similarity between features of all adjacent frames, evaluated using two models: CLIP ViT-L/14~\cite{radford2021learning_clip} and DINOv2~\cite{caron2021emerging_dino}.

\subsection{TGVE}

We also evaluate our PropFly on the TGVE benchmark~\cite{wu2023tgvecvpr}, which contains 76 videos across four editing categories (`style', `object', `background', `multiple'), resulting in a total of 304 editing video-text editing pairs.

For assessing video quality, CLIP temporal consistency, and text-video alignment ($\text{ViCLIP}_{out}$), we use the same settings described in EditVerse~\cite{ju2025editverse}.
Additionally, for text-video direction similarity ($\text{ViCLIP}_{dir}$), we calculate the cosine similarity between the text instruction and the video's directional change from the source to the target.

\section{Ablation Study}
\label{sec:supp_ablation_study}

\subsection{Details of Ablation studies in main paper}
For all ablation studies discussed in the main paper, we trained the models using the same combined dataset (Youtube-VOS~\cite{xu2018youtube} and Pexels~\cite{pexels_website}) for 50,000 iterations, ensuring a fair comparison with our main model.

\vspace{-3mm}
\paragraph{Full Sampling Baseline.}
To implement the `Full Sampling' baseline, we replace our one-step estimation with an iterative solver. 
Given a video sample $\mathbf{x}_0$, noise $\mathbf{x}_1 \sim \mathcal{N}(\mathbf{0},\mathbf{I})$, and a random timestep $t \sim U[0,1]$, we first obtain the intermediate noised latent $\mathbf{x}_t = (1-t)\mathbf{x}_0 + t\mathbf{x}_1$.
Instead of a direct one-step estimation, $\mathbf{x}_t$ is denoised via an ODE solver for $n$ steps, where $n=\lceil N \times t \rceil$ and $N=25$ is the total number of inference steps.
This formula ensures that the solver performs the appropriate number of denoising steps required to traverse the trajectory from the current time $t$ down to $0$.

\begin{figure*}[t!]
    \centering
    \includegraphics[width=\linewidth] {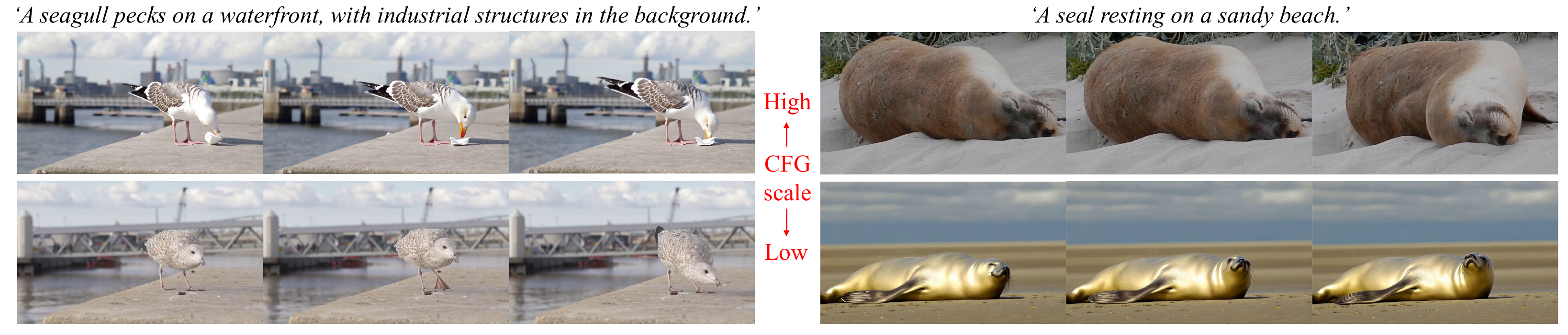}
    \caption{Samples of full sampling with varying CFG scales.}
    \label{fig:full_sampling}
\end{figure*}

We perform this iterative denoising independently using two CFG scales, $\omega_L=1.0$ and $\omega_H=7.0$, yielding the fully sampled latents $\hat{\mathbf{x}}^\text{low}_0$ and $\hat{\mathbf{x}}^\text{high}_0$. 
These are then used as the source and target training pair, with all other settings identical to PropFly.
However, as visualized in Fig.~\ref{fig:full_sampling}, because the two sampling paths are independent, they accumulate numerical errors differently.
Consequently, while the resulting videos align well with their text prompts, they frequently diverge in terms of motion structure, leading to severe motion misalignment between the source and target.

\vspace{-3mm}
\paragraph{Standard FM}
For the `Standard FM' baseline, we train the adapter using the standard flow matching objective rather than our proposed GMFM loss.
Unlike Eq.~\ref{eq:velocity_pred}, we sample new noise \(\mathbf{x}_1' \sim \mathcal{N}(\mathbf{0},\mathbf{I})\) and time step \(t' \sim U(0,1)\), and interpolate the target (edited) latent \(\hat{\mathbf{x}}^\text{high}_{0|t}\) with noise \(\mathbf{x}_1'\), and generate new intermediate noised latent \(\mathbf{x}'_{t'}= (1-t')\hat{\mathbf{x}}^\text{high}_{0|t} + t'\mathbf{x}'_1\).
The adapter predicts the velocity using newly generated noisy latent as below:
\begin{equation}
    \hat{\mathbf{v}}_{\theta, \phi} = {\mathbf{v}}_{\theta, \phi}(\mathbf{x}'_{t'}, t', \mathbf{c}_\text{aug}, \hat{\mathbf{x}}_{0|t}^{\text{low}}, \hat{\mathbf{x}}_{0|t}^{\text{high}}[0]).
\label{eq:abl_fm}
\end{equation}
The adapter is then trained with original flow matching loss as provided in Eq.~\ref{eq:fm_objective}.
As discussed in the main paper, this standard objective guides the model to reconstruct the input video content. 
However, this approach fails to explicitly learn the transformation mapping required to apply the edit from the first frame to the source structure. 
Consequently, the `Standard FM' baseline fails to effectively propagate the edits, often reverting to the original unedited content.

\subsection{CFG variation}

\begin{table}[h!]
    \footnotesize
    \centering
    \caption{Impact of CFG scaling on performance metrics during on-the-Fly data pair generation. Best performance for each metric is highlighted in \textbf{bold}, and the second best performance is indicated by an $\underline{\text{underline}}$.}
    \setlength{\tabcolsep}{3pt}  
\begin{tabular}{cccccc}
\toprule
    
    \(\omega_{H}\) & Pick Score {\small$\uparrow$} & Frame {\small$\uparrow$} & Video {\small$\uparrow$} & CLIP {\small$\uparrow$} & DINO {\small$\uparrow$} \\
     \midrule
    2 & 19.98 & 27.24 & 24.57 & 98.83 & 98.41 \\
    5 & \underline{20.32} & \underline{28.33} & \textbf{25.52} & \underline{99.01} & {98.62} \\
    7 & \textbf{20.35} & \textbf{28.37} & \underline{25.37} & \textbf{99.03} & \underline{98.63} \\
    10 & 20.27 & 28.24 & {25.34} & 99.00 & \textbf{98.80} \\
    20 & 20.19 & \underline{28.33} & 25.27 & 98.85 & {98.62} \\
    \bottomrule
\end{tabular}
    \label{tab:ablation_cfg}
\end{table}

We further analyze the effect of the CFG scale modulation used during our on-the-fly data pair generation. 
As described in the main paper, the semantic gap between the low-CFG ($\omega_L$) and high-CFG ($\omega_H$) latents guides the adapter's learning.
Here, we fix the low scale at $\omega_L = 1.0$ and vary the high scale $\omega_H$ to study its impact on the PropFly model. 
We compare the video editing performance of models trained using different $\omega_H$ values in Table~\ref{tab:ablation_cfg}.

As shown in Table~\ref{tab:ablation_cfg}, we observe that a value of $\omega_H = 7.0$ yields the best overall performance. 
A smaller scale (e.g., $\omega_H = 2.0$) results in lower text alignment, likely because the semantic gap between $\omega_L$ and $\omega_H$ is too small, providing an insufficient supervision signal for the adapter. 
Conversely, a very large scale (e.g., $\omega_H = 20.0$) also leads to a slight drop in video quality (Pick Score). 
This suggests that while the semantic gap is large, the high-CFG latent $\hat{\mathbf{x}}_{0|t}^{\text{high}}$ may begin to contain artifacts or over-saturation, providing a noisy target.
Crucially, the model demonstrates \textbf{robust and high-quality performance across a stable range of $\omega_H$ values} (e.g., 5.0 and 7.0 ). 
This indicates that our method is not sensitive to a specific hyperparameter, but rather succeeds as long as it is provided with a clean, strong semantic signal. Our choice of $\omega_H = 7.0$ simply represents the optimal point within this stable region, providing the best balance of semantic strength and visual quality for supervision.

\subsection{On-the-fly Data Pair Quality}
\label{sec:data_quality}

\begin{figure*}[t!]
    \centering
    \includegraphics[width=\linewidth] {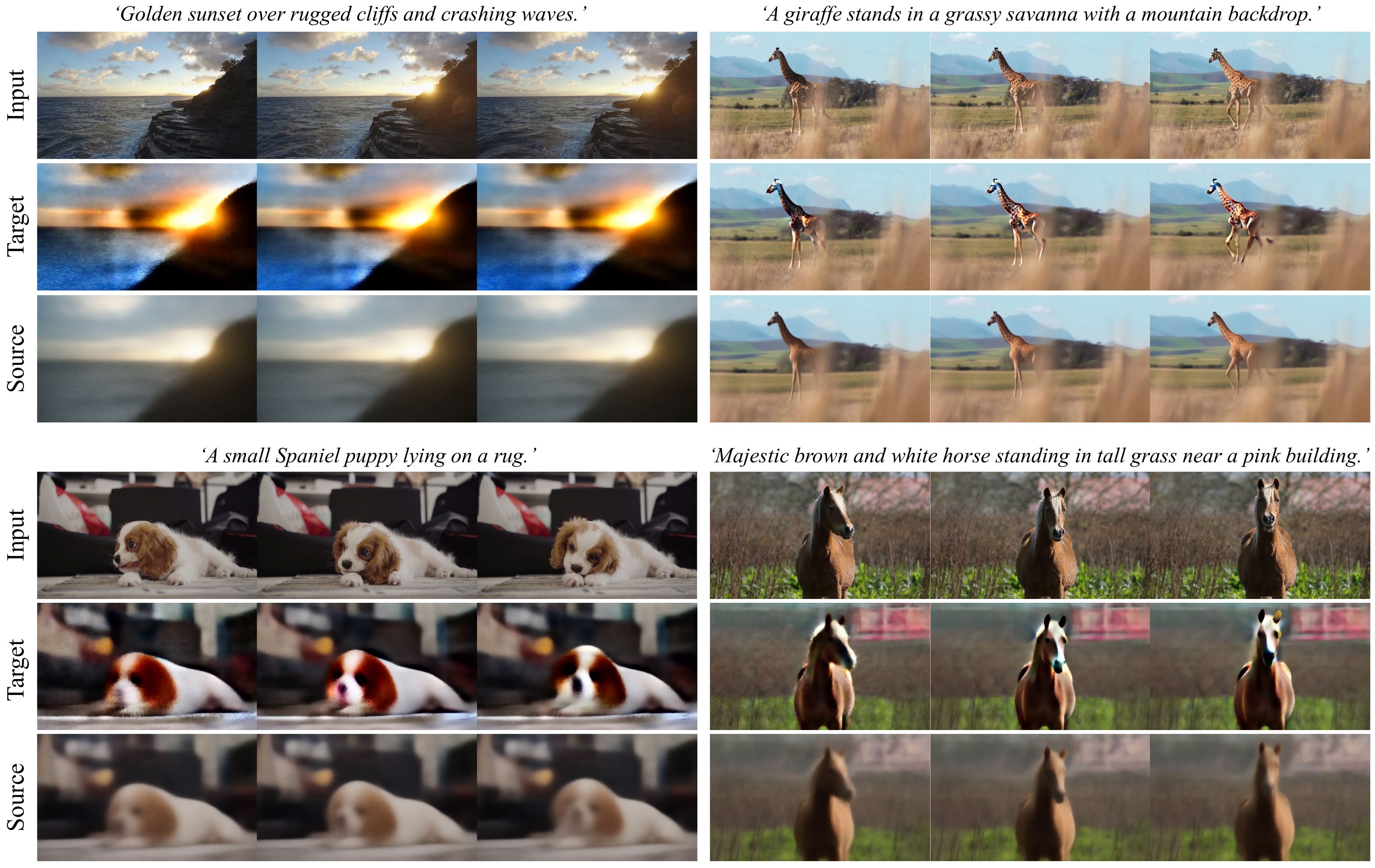}
    \caption{Samples of input videos and their on-the-fly data pairs.}
    \label{fig:onestep_supp}
\end{figure*}

\begin{table}[h!]
    \footnotesize
    \centering
    \caption{Synthetic video data quality comparison. We evaluate the text alignment and motion alignment. $\uparrow$ indicates higher is better.}
    \setlength{\tabcolsep}{3pt}  
\begin{tabular}{lccccc}
\toprule
    
    Target Videos & Text Alignment {\small$\uparrow$} & Motion Alignment {\small$\uparrow$}\\
    \midrule
    Señorita-2M \cite{zi2025senorita} & 20.41 & 78.08 \\
    On-the-fly (Ours) & \textbf{20.97} & \textbf{82.57} \\
    \bottomrule
\end{tabular}
    \label{tab:data_quality}
\end{table}

PropFly relies on synthetic source-target latent pairs generated via the pipeline described in Sec.~\ref{sec:on_the_fly_video_pair}. 
In Fig.~\ref{fig:onestep_supp}, we provide examples of the on-the-fly synthesized data pairs during training.
From the `Input' videos, the `Source' and `Target' videos are synthesized using the one-step clean latent estimation with the low and high CFG values, respectively, during the on-the-fly data pair generation process.
As shown, the transformations between the source and target videos cover the local modification and the global transformation.
Though the visual quality of the on-the-fly data may not seem optimal (as shown in Fig.\ref{fig:onestep_supp}), the difference between the low-CFG and high-CFG samples provides sufficient signal for the model to learn propagation-based video editing, as demonstrated by our video editing results.

To better understand where this ability is coming from, in Table~\ref{tab:data_quality}, we quantitatively assess the quality of decoded video samples produced by our generation process. We focus on two critical dimensions for video editing training:
\begin{itemize}
    \item \textbf{Text Alignment:} Measures how well the target video reflects the style prompt. We compute the average cosine similarity between the synthesized target videos and their corresponding prompts, encoded using ViCLIP-InternVid-10M-Flt~\cite{wang2023internvid_viclip}.
    \item \textbf{Motion Alignment:} Measures how well the motion structure is preserved between the source and target. We utilize the motion fidelity score proposed in STDF~\cite{yatim2024space_stdf}, which averages the correlations between tracklets of the source and edited videos, estimated using CoTracker \cite{karaev2024cotracker}.
\end{itemize}
For comparison, we evaluate 1,000 videos randomly sampled from the `style transfer' subset of the Señorita-2M dataset~\cite{zi2025senorita}, which serves as a representative baseline for offline paired training data.
As shown in the Table~\ref{tab:data_quality}, our on-the-fly generated data exhibits superior text and motion alignment, validating the effectiveness of our CFG-based generation strategy.

\subsection{Generalization to Other Backbones.}

\begin{table}[h!]
    \footnotesize
    \centering
    \caption{Generalization across backbones. We evaluate the generalization capability of PropFly by applying it to both Wan~\cite{wan2025wan_lucy} and LTX-Video~\cite{hacohen2024ltx}, measured on the EditVerseBench-Appearance subset~\cite{ju2025editverse}.}
    \setlength{\tabcolsep}{3pt}  
\begin{tabular}{lccccc}
\toprule
    
    Backbone & Pick Score {\small$\uparrow$} & Frame {\small$\uparrow$} & Video {\small$\uparrow$} & CLIP {\small$\uparrow$} & DINO {\small$\uparrow$} \\
     \midrule
    LTX-2B & 19.87 & 27.37 & 24.68 & 99.02 & 98.61 \\
    Wan-1.3B & 20.35 & 28.37 & 25.37 & 99.03 & 98.63 \\
    Wan-14B & 20.42 & 28.71 & 26.05 & 99.21 & 99.05 \\
    \bottomrule
\end{tabular}
    \label{tab:model_variation}
\end{table}

To demonstrate that PropFly utilizes a model-agnostic training strategy, we apply our method to another pre-trained generative backbone: LTX-Video-2B~\cite{hacohen2024ltx}.
Similar to our main implementation, we attach a VACE adapter to this backbone.
However, since we utilize the pre-trained Image-to-Video (I2V) version of LTX-Video, we do not rely on pre-trained adapter weights (which are typically used to turn T2V models into I2V). Instead, the VACE adapter for LTX-Video is \textbf{trained from scratch}.
As shown in Table~\ref{tab:model_variation}, the LTX-2B model is successfully trained to perform propagation-based global video editing, achieving high-quality results. 
While the absolute performance varies with the inherent strength of each backbone and its corresponding VAE, these results demonstrate that our distillation pipeline can be broadly adopted to equip various video generation models with propagation-based capabilities.

\section{Limitations \& Discussions}
\label{sec:supp_limitations}

While PropFly demonstrates robust performance in propagation-based video editing, it is subject to certain limitations. First, since our method leverages the generative priors of a pre-trained T2V backbone, the resulting video quality and motion dynamics are naturally influenced by the native generative capacity of the base model.
It should be noted that this also provides a practical scalability advantage. As stronger T2V backbones are developed, our PropFly can directly benefit from them simply by replacing the underlying model, without modifying the major propagation pipeline itself. 

Second, although our approach eliminates the need for costly offline dataset construction, the on-the-fly data pair generation introduces a modest computational overhead during training due to the additional sampling steps. 
Note that this pipeline allows the supervision distribution to be adapted to changes in the backbone or prompt design, while maintaining the inference-time efficiency of the overall framework. 

Finally, our current training pipeline relies on descriptive text guidance (e.g., ``A panda is walking'') rather than direct edit instructions (e.g., ``Change the bear to a panda''). 
Although this limits the ability to perform edits based purely on text instructions (e.g., instructive V2V), it allows the model to leverage strong visual guidance from the edited first frame. This trade-off results in superior control over content preservation and temporal consistency in the video propagation setting.

Despite these limitations, we believe our PropFly provides a simple and scalable foundation for propagation-based video editing without paired training data, and can serve as a strong basis for more general video editing frameworks.

\section{Further Qualitative Comparison}
\label{sec:supp_qualitative_comparison}

We provide extensive additional qualitative results to further demonstrate the capabilities of PropFly.
In addition to the figures presented in this document, we provide video results to demonstrate temporal consistency. 
These can be viewed in the \href{./Supplementary_Videos/comparison_videos.html}{propagation\_comparison.html} and \href{./Supplementary_Videos/PropFly_videos.html}{PropFly\_videos.html} files located in the \textit{\textbf{Supplementary\_Videos}} directory. 

\subsection{Qualitative Comparison on DAVIS}

\begin{figure*}[h!]
    \centering
    \includegraphics[width=\linewidth] {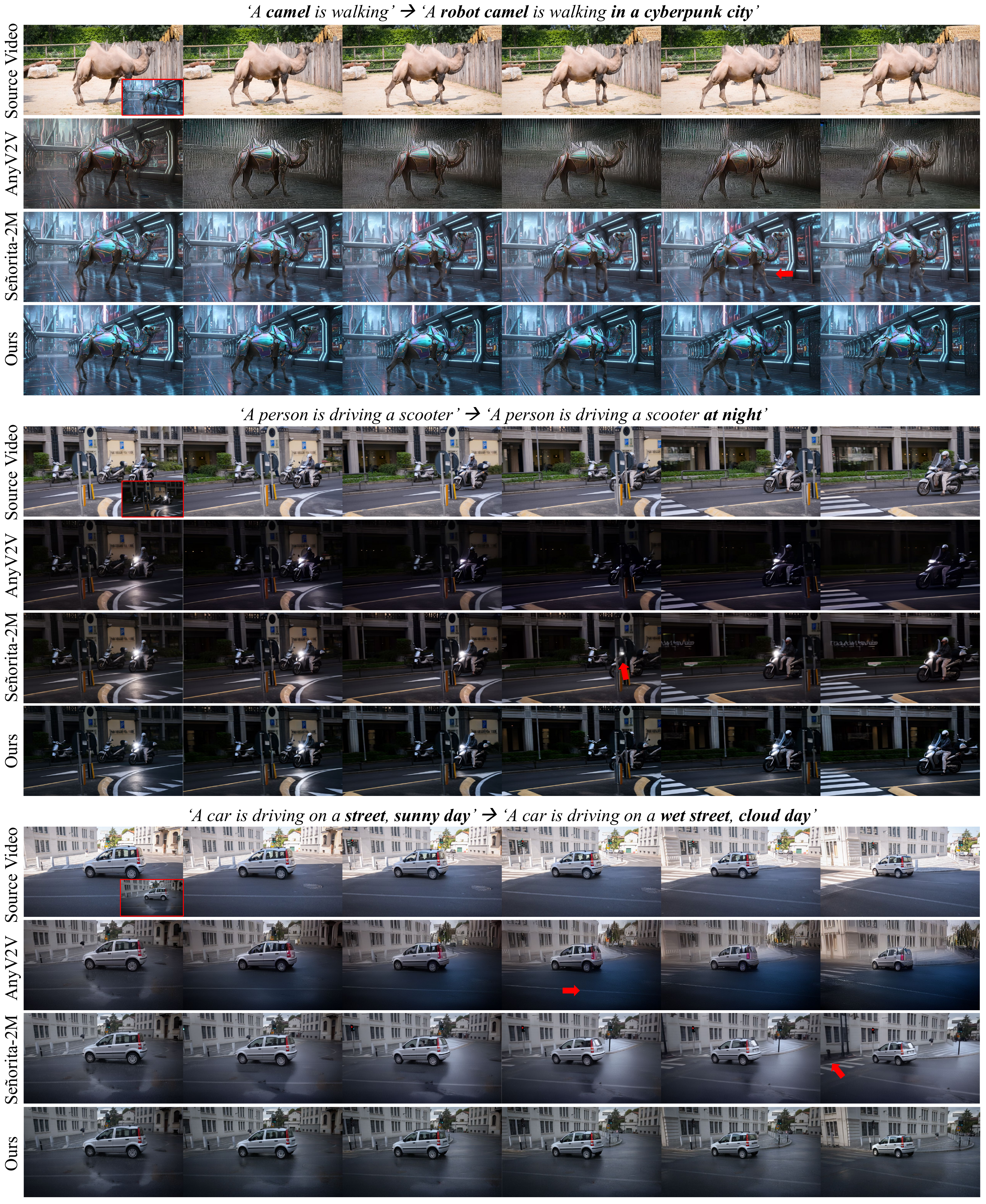}
    \caption{Video quality comparison between propagation-based video editing methods on the DAVIS dataset~\cite{Perazzi2016}.}
    \label{fig:qualitative_davis_supp1}
\end{figure*}

\begin{figure*}[h!]
    \centering
    \includegraphics[width=\linewidth] {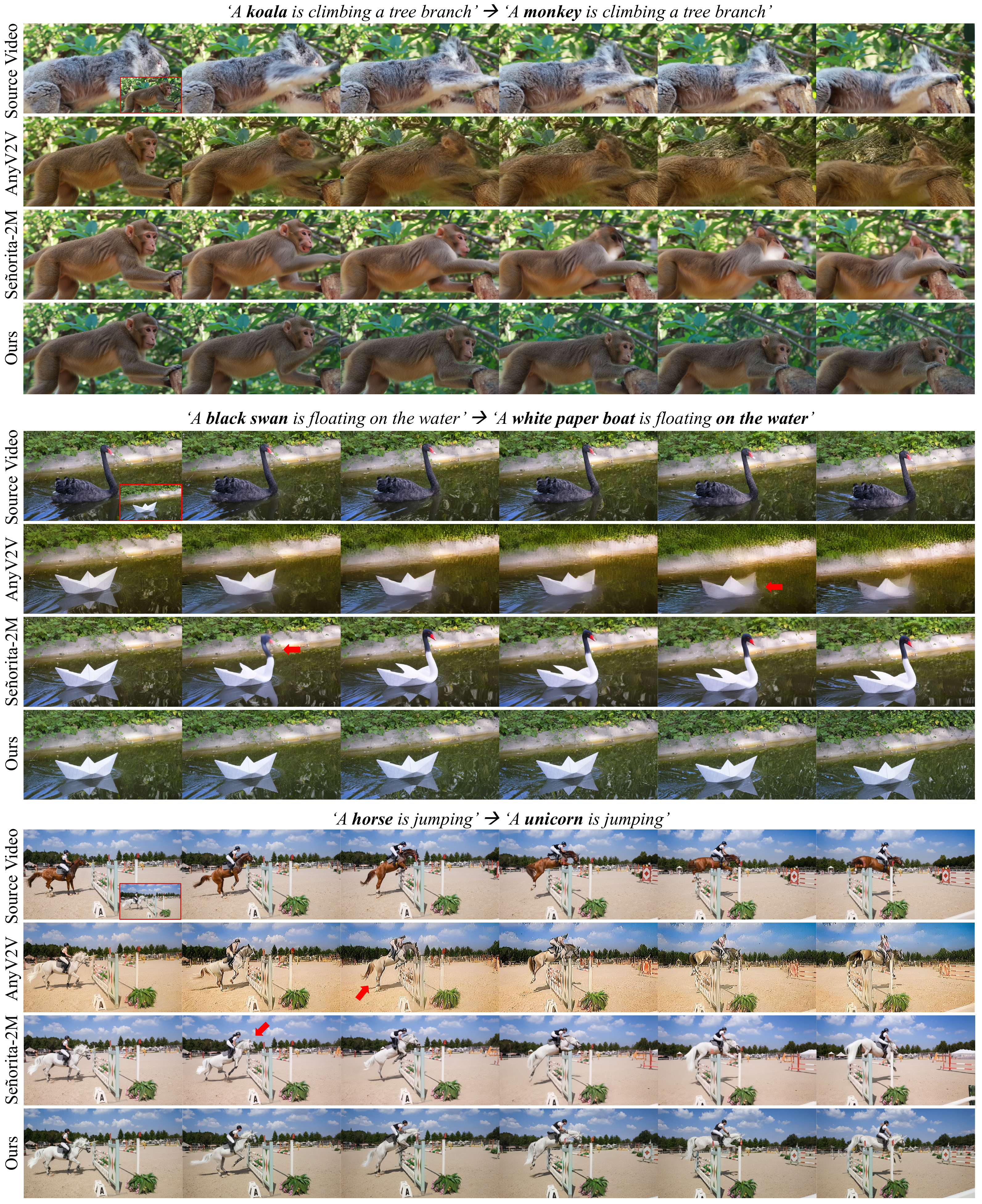}
    \caption{Video quality comparison between propagation-based video editing methods on the DAVIS dataset~\cite{Perazzi2016}.}
    \label{fig:qualitative_davis_supp2}
\end{figure*}

In Figs.~\ref{fig:qualitative_davis_supp1} and~\ref{fig:qualitative_davis_supp2}, we present a visual comparison on the DAVIS dataset~\cite{Perazzi2016}, contrasting PropFly with leading propagation-based methods: AnyV2V \cite{ku2024anyv2v} and Señorita-2M \cite{zi2025senorita}.
As shown, AnyV2V often struggles to perform complex edits that require transforming both the object and the background simultaneously.
Similarly, Señorita-2M \cite{zi2025senorita} frequently fails to propagate the transformation consistently or preserve the context of the original videos.
In contrast, our PropFly generates high-fidelity videos that faithfully propagate the target transformation while preserving the integrity of the source context.

\subsection{Qualitative Comparison on EditVerseBench}
We also provide comprehensive qualitative comparisons on the EditVerseBench \cite{ju2025editverse}. 
We compare against a wide range of baselines, including InsV2V \cite{cheng2023consistent_insv2v}, LucyEdit \cite{wan2025wan_lucy}, STDF \cite{yatim2024space_stdf}, VACE \cite{vace}, EditVerse \cite{ju2025editverse}, Runway \cite{RunwayAleph2025}, and Señorita-2M \cite{zi2025senorita}, utilizing their publicly available results.

\begin{figure*}[h!]
    \centering
    \includegraphics[width=\linewidth] {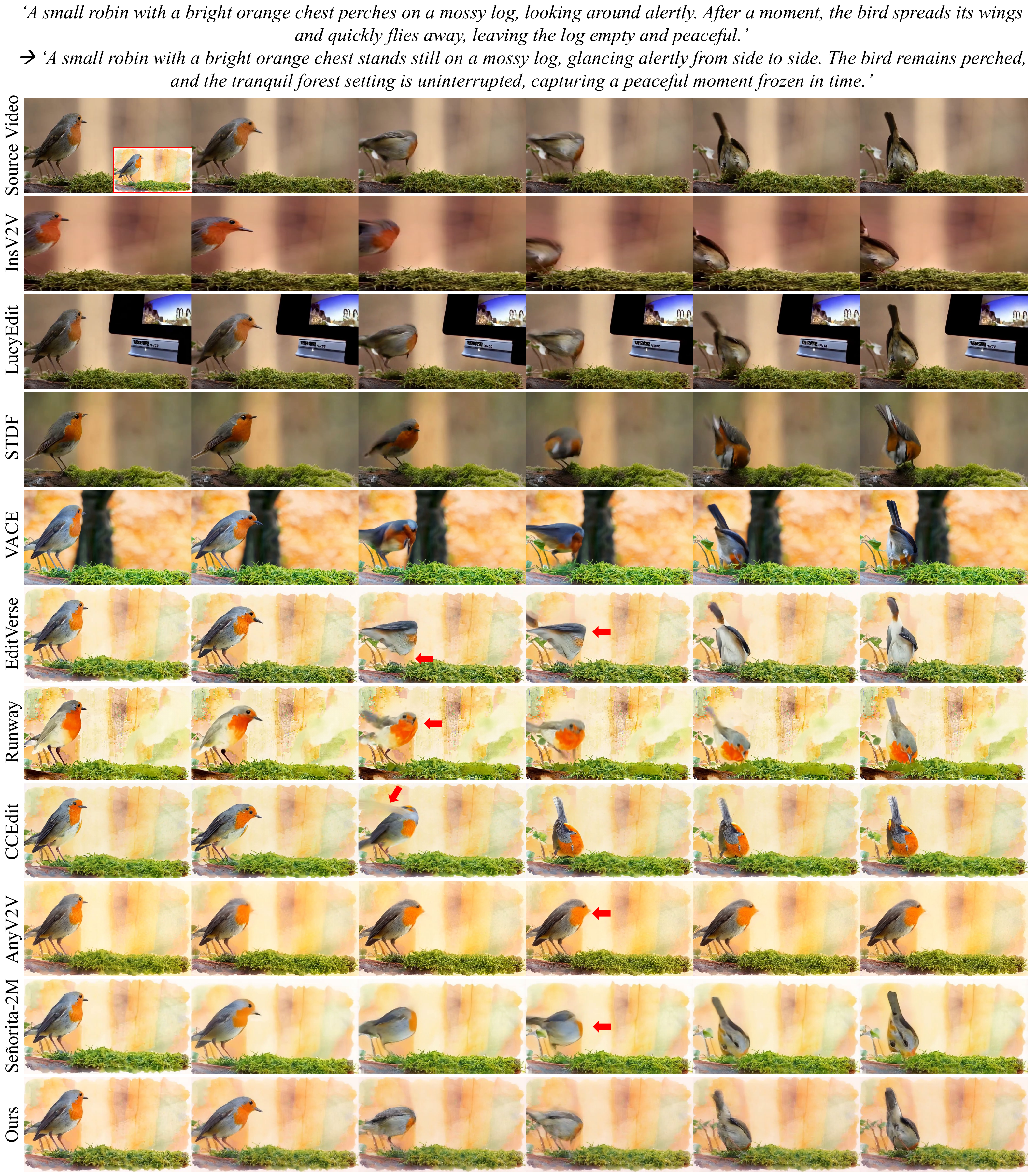}
    \caption{Video quality comparison on the \textbf{Propagation} task in EditVerseBench \cite{ju2025editverse}}
    \label{fig:qualitative_ev_1}
\end{figure*}

\begin{figure*}[h!]
    \centering
    \includegraphics[width=\linewidth] {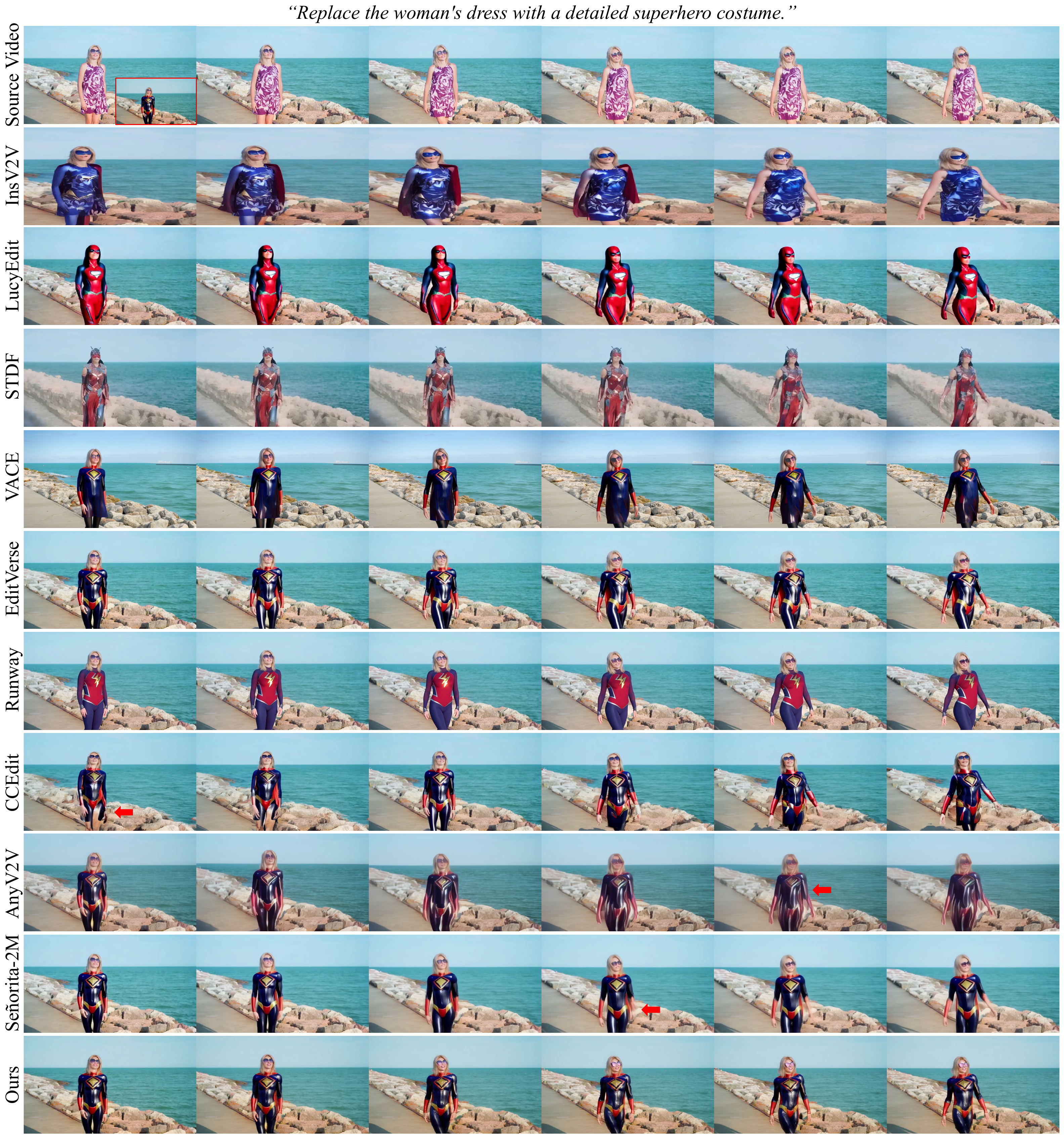}
    \caption{Video quality comparison on the \textbf{Change object} task in EditVerseBench \cite{ju2025editverse}}
    \label{fig:qualitative_ev_2}
\end{figure*}

\begin{figure*}[h!]
    \centering
    \includegraphics[width=\linewidth] {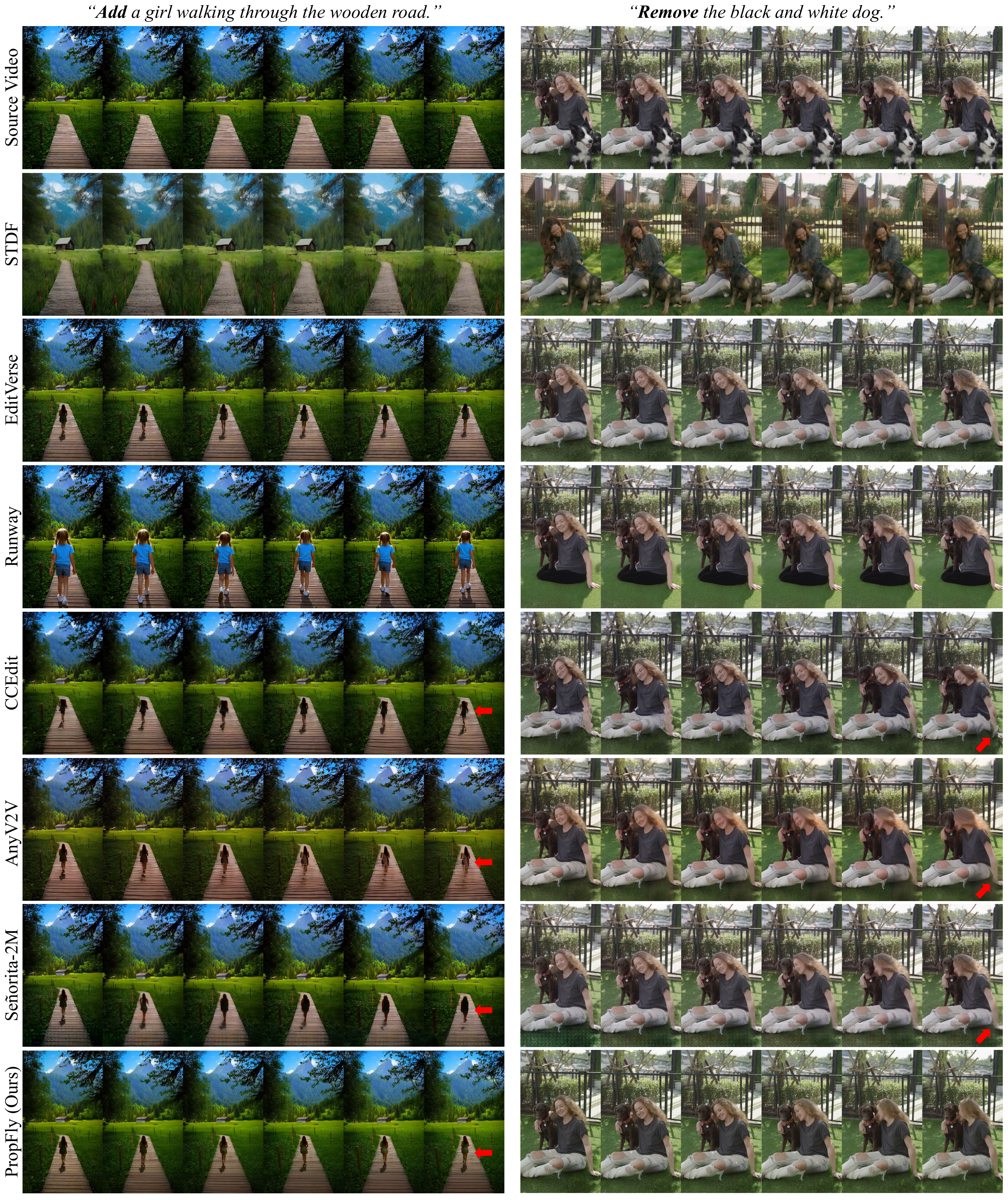}
    \caption{Video quality comparison on the \textbf{Add object} and \textbf{Remove object} tasks in EditVerseBench \cite{ju2025editverse}}
    \label{fig:qualitative_ev_3}
\end{figure*}

To ensure a fair comparison, we adopt specific protocols for the input conditions depending on the task type.
For the propagation tasks shown in Fig.~\ref{fig:qualitative_ev_1}, we employ the \textit{same} provided edited first frame for PropFly and other propagation-based baselines~\cite{ku2024anyv2v, zi2025senorita}. 
One exception is CCEdit~\cite{feng2024ccedit}, where we utilize the edited \textit{center} frame from the EditVerse baseline to accommodate its bidirectional propagation mechanism.
For general editing tasks (Figs.~\ref{fig:qualitative_ev_2} and~\ref{fig:qualitative_ev_3}), such as object modification, addition, or removal, we utilize the edited first frame generated by EditVerse~\cite{ju2025editverse} as the starting condition for our propagation.

Fig.~\ref{fig:qualitative_ev_1} shows the video propagation comparison. 
We observe that text-guided video editing methods such as STDF~\cite{yatim2024space_stdf}, LucyEdit~\cite{wan2025wan_lucy}, and InsV2V~\cite{cheng2023consistent_insv2v} fail to propagate the specific transformed style, as they rely solely on text instructions.
Other methods~\cite{vace, ju2025editverse, RunwayAleph2025, feng2024ccedit, ku2024anyv2v, zi2025senorita} demonstrate propagation capabilities, they often struggle to reconstruct complex dynamics, such as the fast motion of the bird.
In contrast, our PropFly successfully propagates the transformed style across the entire video while faithfully preserving the original motion.
In Fig.~\ref{fig:qualitative_ev_2}, other propagation methods~\cite{feng2024ccedit, ku2024anyv2v, zi2025senorita} often fail to consistently propagate the woman's changed clothes or accurately reconstruct her motion. 
On the other hand, our PropFly robustly maintains the edited appearance throughout the sequence and successfully preserves the fidelity of the subject's movement.

We also compare the object addition and removal quality across methods in Fig.~\ref{fig:qualitative_ev_3}.
For object addition (left side of Fig.~\ref{fig:qualitative_ev_3}), other propagation-based methods~\cite{feng2024ccedit, ku2024anyv2v, zi2025senorita} struggle to synthesize the girl's motion naturally, whereas our PropFly generates faithful and coherent motion.
For object removal (right side of Fig.~\ref{fig:qualitative_ev_3}), while other baselines~\cite{feng2024ccedit, ku2024anyv2v, zi2025senorita} fail to plausibly fill the removed region, our PropFly effectively synthesizes the girl's left hand, maintaining temporal consistency.
It is worth noting that PropFly is not explicitly trained for object addition or removal tasks, nor does it utilize mask guidance. 
However, these results demonstrate that our model learns a sufficiently robust and generalized transformation to handle such complex structural edits effectively.

\end{document}